\documentclass{article}
\usepackage{PRIMEarxiv}

\usepackage[utf8]{inputenc} 
\usepackage[T1]{fontenc}    
\usepackage{url}    
\usepackage{booktabs}      
\usepackage{amsfonts}      
\usepackage{nicefrac}       
\usepackage{microtype}      
\usepackage{fancyhdr}       
\usepackage{amsmath}          
\usepackage{graphicx}          
\usepackage{xcolor}
\usepackage{amsthm}           
\usepackage{mathtools}
\usepackage{lastpage}          
\usepackage{enumerate}         
\usepackage{parskip}          
\usepackage{caption}
\usepackage{listings}          
\usepackage{algorithm}
\usepackage{algorithmic}
\usepackage[symbol]{footmisc}

\usepackage[round]{natbib}

\newcommand{\E}{\mathbb{E} \,} % Expected Value
\newcommand{\V}{\mathbb{V}} % pseudo-variance operator 
\newcommand{\grad}{\nabla} % Gradient
\newcommand{\N}{\mathcal{N}} % Script N for Normal Distribution
\newcommand{\commentout}[1]{}

\DeclareMathOperator{\Cov}{Cov} % Covariance 
\DeclareMathOperator{\Var}{Var} % Variance
\DeclareMathOperator{\tr}{tr} % Trace
\newtheorem{theorem}{Theorem}[section]

\newtheorem{lemma}[theorem]{Lemma}
\theoremstyle{definition}

\newtheorem{assumption}{Assumption}
\theoremstyle{remark}

\title{Preferential Subsampling for Stochastic Gradient Langevin Dynamics}

\author{
  Srshti Putcha\thanks{Correspondence: \texttt{srshti.putcha@gmail.com}} \\
  STOR-i Centre for Doctoral Training \\
  Lancaster University 
   \And
  Christopher Nemeth \\
  Department of Mathematics and Statistics \\
  Lancaster University
  \And
  Paul Fearnhead \\
  Department of Mathematics and Statistics \\
  Lancaster University 
}
\begin{document}
\maketitle
%%%%%%%%%%%%%%%%%%%%%%%%%%%%%%%%%%%%%%%%%%%%%%%%%%%%%%%%%%%
\vspace{-15pt}
\begin{abstract}
Stochastic gradient MCMC (SGMCMC) offers a scalable alternative to traditional MCMC, by constructing an unbiased estimate of the gradient of the log-posterior with a small, uniformly-weighted subsample of the data. While efficient to compute, the resulting gradient estimator may exhibit a high variance and impact sampler performance. The problem of variance control has been traditionally addressed by constructing a better stochastic gradient estimator, often using control variates. We propose to use a discrete, non-uniform probability distribution to preferentially subsample data points that have a greater impact on the stochastic gradient. In addition, we present a method of adaptively adjusting the subsample size at each iteration of the algorithm, so that we increase the subsample size in areas of the sample space where the gradient is harder to estimate. We demonstrate that such an approach can maintain the same level of accuracy while substantially reducing the average subsample size that is used.
\end{abstract}
% keywords can be removed
%%%%%%%%%%%%%%%%%%%%%%%%%%%%%%%%%%%%%%%%%%%%%%%%%%%%%%%%%%%
\section{Introduction}
\label{sec:intro}
Markov chain Monte Carlo (MCMC) algorithms are a popular family of methods to conduct Bayesian
inference. Unfortunately, running MCMC on large
datasets is generally  computationally expensive, which often limits the use of MCMC by practitioners. The Metropolis-Hastings  algorithm \citep{metropolis1953equation, hastings1970monte}, in particular, requires a scan
of the full dataset at each iteration to calculate the acceptance probability. 

Stochastic gradient Markov chain Monte Carlo (SGMCMC) algorithms are a family of scalable methods, which aim to address this issue \citep{welling2011bayesian, nemeth2021stochastic}. These algorithms aim to leverage the efficiency of gradient-based MCMC proposals \citep{roberts1996exponential, neal2011mcmc}. They  reduce the per-iteration computational cost by constructing an unbiased, noisy estimate of the gradient of the log-posterior, using only a small data subsample. In this paper, we focus on samplers that rely on the overdamped Langevin diffusion \citep{roberts1996exponential}, however, our proposed methodology can be applied more generally to other SGMCMC algorithms, such as those based on Hamiltonian dynamics \citep{chen2014}. 

The high variance inherently present in the stochastic gradient estimator can degrade sampler performance and lead to poor convergence. As such, variance control has become an important area of research within the SGMCMC literature, and is often required to make these algorithms practical \citep{dubey2016variance, chatterji2018theory, baker2019control, chen2019convergence}. 

In this paper, we propose a new method designed to reduce the variance in the stochastic gradient. We use a discrete, non-uniform probability distribution to preferentially subsample data points and to re-weight the stochastic gradient. In addition, we present a method for adaptively adjusting the size of the subsample chosen at each iteration. 

%%%%%%%%%%%%%%%%%%%%%%%%%%%%%%%%%%%%%%%%%%%%%%%%%%%%%%%%%%%%%%
\section{Stochastic gradient MCMC}
\label{sec:sgmcmc}
Let $\theta \in \mathbb{R}^d$ be a parameter vector and denote independent observations $\mathbf{x} = \{x_i\}_{i=1}^N$ ($N \gg 1$). The probability density of the $i$-th  observation, given parameter $\theta$, is $p(x_i | \theta)$ and the prior density for the parameters is $p(\theta)$. In a Bayesian context, the target of interest is the posterior density, $\pi(\theta) \vcentcolon= p(\theta | \mathbf{x}) \propto p(\theta) \prod_{i=1}^N p(x_i| \theta).$

For convenience, we define $f_i(\theta) = - \log p(x_i | \theta)$ for $i=1,\ldots, N$, with $f_0(\theta) = - \log p(\theta)$ and $f(\theta) = f_0(\theta) + \sum_{i=1}^N f_i(\theta)$. In this setting, the posterior density can be rewritten as, $\pi(\theta) \propto \exp(-f(\theta))$. \par

\subsection{The Langevin diffusion}
The Langevin diffusion, $\theta(t)$, is defined by the stochastic differential equation,
\begin{align} \label{eq:old}
    d\theta(t) = -\frac{1}{2}\nabla f(\theta(t))dt + dB_t,
\end{align} where $\nabla f(\theta(t))dt$ is a drift term and $B_t$ denotes a $d$-dimensional Wiener process. Under certain regularity conditions, the stationary distribution of this diffusion is the posterior $\pi$ \citep{roberts1996exponential}.  In practice, we need to discretise Eq.~\eqref{eq:old} in order to simulate from it and this introduces error. For a small step-size $\epsilon > 0$, the Langevin diffusion can be approximated by
\begin{align} \label{eq:ula}
    \theta^{(t+1)} = \theta^{(t)} -\frac{\epsilon}{2}\nabla f\big(\theta^{(t)}\big) + \sqrt{\epsilon}\,\eta^{(t)},
    \end{align}
where the noise $\eta^{(t)} \sim \mathcal{N}_d(\mathbf{0}, I_{d\times d})$ is drawn independently at each update. The dynamics implied by Eq.~\eqref{eq:ula} provide a simple way to sample from the Langevin diffusion. The level of discretisation error in the approximation is controlled by the size of $\epsilon$ and we can achieve any required degree of accuracy if we choose $\epsilon$ small enough.

The unadjusted Langevin algorithm (ULA) \citep{parisi1981} is a simple sampler that simulates from Eq.~\eqref{eq:ula} but does not use a Metropolis-Hastings correction \citep{metropolis1953equation, hastings1970monte}. Thus, the samples obtained from ULA produce a biased approximation of $\pi$. The per-iteration computational cost of ULA is smaller than that of the Metropolis-adjusted Langevin algorithm \citep{roberts1998optimal} due to the removal of the Metropolis-Hastings step. However, the computational bottleneck for ULA lies in the  $O(N)$ calculation of the full data gradient $\nabla f\big(\theta^{(t)}\big)=\nabla f_0\big(\theta^{(t)}\big) + \sum_{i=1}^N \nabla f_i \big(\theta^{(t)}\big)$
at every iteration. This calculation can be problematic if $N$ is large. 

\subsection{Stochastic gradient Langevin dynamics}
The stochastic gradient Langevin dynamics (SGLD) algorithm attempts to improve the per-iteration computational burden of ULA by replacing the full-data gradient with an unbiased estimate \citep{welling2011bayesian}. Let the full-data gradient of $f(\theta)$ be given by 
$$g^{(t)} = \nabla f\big(\theta^{(t)}\big)=\nabla f_0\big(\theta^{(t)}\big) + \sum_{i=1}^N \nabla f_i \big(\theta^{(t)}\big).$$
The  unbiased estimate of $g^{(t)}$ proposed by \citet{welling2011bayesian} takes the form
\begin{align}
\label{eq:ghat}
\hat{g}^{(t)} = \nabla f_0\big(\theta^{(t)}\big) + \frac{N}{n} \sum_{i \in \mathcal{S}^t} \nabla f_i \big(\theta^{(t)}\big),
\end{align}
where $\mathcal{S}^t$ is a subset of $\{1, \ldots, N\}$ and $|\mathcal{S}^t| = n$ ($n \ll N$) is the subsample size. A single update of SGLD is thus given by,
\begin{align} \label{eq:sgldupdate}
  \theta^{(t+1)} \leftarrow \theta^{(t)} - \frac{\epsilon^{(t)}}{2} \cdot \hat{g}^{(t)}  + \xi^{(t)},
\end{align} 
where $\xi^{(t)} \sim \mathcal{N}_d(0, \epsilon^{(t)} I_{d \times d})$ and $
\{\epsilon^{(t)}\}$ corresponds to a schedule of step-sizes which may be fixed \citep{vollmer2016non} or decreasing \citep{teh2016consistency}. The full SGLD pseudocode is provided in Algorithm~\ref{alg:sgld}.

\begin{algorithm}
  \caption{SGLD} \label{alg:sgld}
  \begin{algorithmic}[1]
    \STATE{Input: initialise $\theta^{(1)}$, batch size $n$, step-sizes  $\{\epsilon^{(t)}\}$.}
   \FOR{$t = 1, 2, \ldots, T$}
   \STATE{Sample indices $S^t \subset \{1,\ldots,N\}$ with or without replacement.}
   \STATE{Calculate $\hat{g}^{(t)}$ using Eq.~\eqref{eq:ghat}.}
   \STATE{Update parameters according to Eq.~\eqref{eq:sgldupdate}. }
    \ENDFOR 
    \RETURN $\theta^{(T+1)}$ 
 \end{algorithmic}
\end{algorithm}
\citet{welling2011bayesian} note that if the step-size $\epsilon^{(t)} \rightarrow 0$ as $t \rightarrow \infty$, then the Gaussian noise (generated by $\xi^{(t)}$) dominates the noise in the stochastic gradient term. For large $t$, the algorithm approximately samples from the posterior using an increasingly accurate discretisation of the Langevin diffusion. In practice, SGLD does not mix well when the step-size is decreased to zero and so a small fixed step-size $\epsilon$ is typically used instead.

\subsection{Control variates for SGLD}
The naive stochastic gradient proposed by \citet{welling2011bayesian} may exhibit a relatively high variance for small subsamples of data. The more faithful a stochastic gradient estimator is to the full-data gradient, the better we can expect SGLD to perform. Therefore, it is natural to consider alternatives to the estimator given in Eq.~\eqref{eq:ghat} which minimise the variance. 

Let $\hat{\theta}$ be a fixed value of the parameter, typically chosen to be close to the mode of the target posterior density.  The control variate gradient estimator proposed by \citet{baker2019control} takes the form,
\begin{align} \label{eq:ghat_cv}
    \hat{g}_{cv}^{(t)} = \big[\nabla f(\hat{\theta}) + \nabla f_0\big(\theta^{(t)}\big) - \nabla f_0(\hat{\theta})\big] \, + \,\\  \frac{N}{n} \sum_{i \in \mathcal{S}^t} \big[\nabla f_i \big(\theta^{(t)}\big) - \nabla f_i(\hat{\theta}) \big]. \nonumber
\end{align} When $\theta^{(t)}$ is close to $\hat{\theta}$, the variance of the gradient estimator will be small. This is shown formally in Lemma 1 of \cite{baker2019control}.  

The SGLD-CV algorithm is the same as SGLD given in Algorithm \ref{alg:sgld}, except with $\hat{g}_{cv}^{(t)}$ substituted in place of $\hat{g}^{(t)}$. Implementing the SGLD-CV estimator involves a one-off pre-processing step to find $\hat{\theta}$, which is typically done using stochastic gradient descent (SGD) \citep{bottou2018, baker2019control}. The gradient terms $\nabla f_i \big(\hat{\theta})$ are calculated and stored. While these steps are both $O(N)$ in computational cost, the optimisation step to find the mode can replace the typical burn-in phase of the SGLD chain. The MCMC chain can then be initialised at the posterior mode itself. The full pseudocode for SGLD-CV is provided in Algorithm 3 within Appendix \ref{sec:alg-section}.

%%%%%%%%%%%%%%%%%%%%%%%%%%%%%%%%%%%%%%%%%%%%%%%%%%%%%%%%%%%%%%%%%%%%%%%%%%%%%%%%%%%% 
\section{Preferential data subsampling}
\label{sec:method}
Let $\mathcal{S}$ be a subsample of size $n$ generated with replacement, such that the probability that the $i$-th data point, $x_i$, appears in $\mathcal{S}$ is $p_i$. The expected number of times $x_i$ is drawn from the dataset is $np_i$. A standard implementation of SGLD would assume uniform subsampling, i.e. that $p_i = \frac{1}{N}$ for all $i$. However, if the observations vary in their information about the parameters, then assigning a larger probability to the more informative observations would be advantageous. Preferential subsampling assigns a strictly-positive, user-chosen weight to each data point, such that we minimise the variance of the estimator of the gradient. 

We need to construct a discrete probability distribution $\mathbf{p}^{(t)} = (p_{1}^t, \ldots, p_{N}^t)^T$ (where $p_i^{(t)} > 0$ for all $i$ and $\sum_i p_{i}^t = 1$) that can be used to draw subsamples of size $n$ at each iteration $t$ and to reweight the stochastic gradient accordingly. If $\mathbf{p}^{(t)}$ is time-invariant (i.e. $p_i^{(t)} = p_i$ for all $i$), the preferential subsampling scheme is static. Otherwise, the  subsampling weights will be dynamic or state-dependent. 

For a given stochastic gradient $\tilde{g}$, the noise term associated with $\tilde{g}$ is given by $\xi^{(t)} = \tilde{g}^{(t)} - g^{(t)}$. Taking expectations over $\mathbf{p}^{(t)}$, a simple scalar summary of the variance of the noise $\xi^{(t)}$ can be found by evaluating: 
\begin{align} \label{eq:pseudovar}
 \E \bigg( \big\|\xi^{(t)}\big\|^2 \bigg) = \tr\bigg(\Cov \big(\tilde{g}^{(t)}\big)\bigg). 
\end{align} We will refer to Eq.~\eqref{eq:pseudovar} as the \textit{pseudo-variance}\footnote{See Appendix \ref{sec:pseudovar} for a full derivation of the pseudo-variance.} of $\tilde{g}^{(t)}$,  $\V\big(\tilde{g}^{(t)}\big) $, from now on. We intend to use the pseudo-variance as a proxy for the variance of the stochastic gradient.\footnote{Note that $\tilde{g}$ is a $d$-dimensional random vector (where typically $d>1$). Eq.~\eqref{eq:pseudovar} is the sum of the variances of the elements of $\tilde{g}$. The term ``pseudo-variance" allows us to easily distinguish between Eq.~\eqref{eq:pseudovar} and the variance-covariance matrix of $\tilde{g}$.} In all further analysis, $\|\cdot\|$ refers to the Euclidean norm. \par

In order to minimise the pseudo-variance, we need to find a preferential subsampling distribution $\mathbf{p}^*$ which minimises the following problem: 
 \begin{align} \label{eq:min1}
   \min_{\mathbf{p}^{(t)}, \, p_{i}^t \in [0,1], \sum_i p_{i}^t = 1} \mathbb{V}\big(\tilde{g}^{(t)} \big). 
 \end{align} 

Existing non-asymptotic convergence results for SGLD-type methods\footnote{Theorem 4 of \citet{dalalyan2019} and Theorem 1 of \citet{baker2019control} are two such examples.} demonstrate the importance in controlling the variance of the stochastic gradient. These results give the error of SGLD in terms of bounds on the (bias and) variance of the estimator of the gradient of the log posterior. Therefore, constructing a better stochastic gradient  estimator - for instance via preferential subsampling - will lead to a reduction in the error bound of the underlying SGLD method. 

\subsection{SGLD with preferential subsampling}  \label{sec:method_ps}
An alternative gradient estimator for SGLD can be given by reweighting the simple estimator defined in Eq.~\eqref{eq:ghat} \citep{welling2011bayesian},
\begin{align} \label{eq:gtilde}
  \tilde{g}^{(t)} =\nabla f_0\big(\theta^{(t)}\big) +  \frac{1}{n} \sum_{i \in \mathcal{S}^t} \frac{1}{p_{i}^t}\nabla f_i \big(\theta^{(t)}\big),
 \end{align} where $S^t \subset \{1, \ldots, N\}$ is selected according to $\mathbf{p}^{(t)}$ and $|S^t| = n \enspace (n \ll N)$. The pseudocode for the SGLD with preferential subsampling (SGLD-PS) algorithm is outlined in Algorithm 4 within Appendix \ref{sec:alg-section}. 

As we correct for the non-uniform subsampling of data points by reweighting each gradient term, it follows that the stochastic gradient estimator given in Eq.~\eqref{eq:gtilde} is unbiased. This is synonymous with the standard properties of importance sampling estimators \citep{robert2004}. We note that there is an extra $O(n)$ computational cost associated with reweighting the stochastic gradient in this manner at each iteration.

The following result obtains the optimal solution to Problem~\eqref{eq:min1}. \begin{lemma} \label{lemma3.1}
For the unbiased SGLD-PS gradient estimator in Eq.~\eqref{eq:gtilde}, minimising  Problem~\eqref{eq:min1} is equivalent to minimising the following
\begin{align} \label{eq:min2}
  \min_{\mathbf{p}^{(t)}}  \frac{1}{n}\sum_{i=1}^N \frac{1}{p_{i}^t} \big\|\nabla f_i \big(\theta^{(t)}\big) \big\|^2.
\end{align}
The optimal weights which minimise the pseudo-variance are thus given by
\begin{align}  \label{eq:optp1} p_{i}^{t} = \frac{\| \nabla f_i \big(\theta^{(t)}\big)\|}{\sum_{k=1}^N \| \nabla f_k \big(\theta^{(t)}\big)  \| }\, \text{ for } i=1, \ldots, N. 
\end{align}
\end{lemma}

Although a solution to Eq.~\eqref{eq:min2} can be found, the resulting sampling algorithm would not be practical, as the optimal weights depend on the current state $\theta^{(t)}$. Therefore the subsampling distribution given in Eq.~\eqref{eq:optp1} requires $N$ gradient calculations per iteration. For large datasets, these weights would be very expensive to store and calculate at each iteration, making the algorithm impractical. 

We can instead approximate the optimal weights given in Eq.~\eqref{eq:optp1}, such that they are not state-dependent and therefore do not need to be updated at each iteration. These approximate weights can be calculated as an initial pre-processing step before the main sampling algorithm is run. 

A fairly simple approximation of the optimal weights given in Eq.~\eqref{eq:optp1} for SGLD-PS would require substituting the current state $\theta^{(t)}$ with some alternative fixed point. The posterior mode, $\hat{\theta}$, is a sensible choice as it represents the most probable estimate of the parameters in a Bayesian paradigm. In this case, the approximate subsampling scheme would be given by, 
\begin{align}  \label{eq:approxp1} p_{i} = \frac{\| \nabla f_i \big(\hat{\theta}\big)\|}{\sum_{k=1}^N \| \nabla f_k \big(\hat{\theta}\big)  \| }\,\text{ for } i=1, \ldots, N. 
\end{align} As with SGLD-CV, the posterior mode could be estimated using SGD and the MCMC chain could then be initialised at the posterior mode. 

In practice, the subsampling weights in Eq.~\eqref{eq:approxp1} are calculated only once with an $O(N)$ preprocessing step and then used statically (i.e without update). The resulting SGLD-PS algorithm would calculate the stochastic gradient given in Eq.~\eqref{eq:gtilde} using these fixed weights. 
\subsection{SGLD-CV with preferential subsampling}
\label{sec:method_cv}
The control-variates gradient estimator can be modified to accommodate a preferential subsampling scheme in a similar manner. In this case, we would obtain
\begin{align} \label{eq:gtilde_cv}
  \tilde{g}^{(t)} = \big[\nabla f(\hat{\theta}) + \nabla f_0\big(\theta^{(t)}\big) - \nabla f_0(\hat{\theta})\big]\, + \, \\ \frac{1}{n} \sum_{i \in \mathcal{S}^t} \frac{1}{p_{i}^t} \big[\nabla f_i \big(\theta^{(t)}\big) - \nabla f_i(\hat{\theta}) \big] \nonumber.
 \end{align}  The pseudocode for the modified SGLD-CV algorithm (SGLD-CV-PS) is given in Algorithm 5 within Appendix \ref{sec:alg-section}.
The following result provides the optimal solution to Problem~\eqref{eq:min1}. 
 \begin{lemma} 
\label{lemma3.2}
For the unbiased SGLD-CV-PS gradient estimator in Eq.~\eqref{eq:gtilde_cv}, minimising  Problem~\eqref{eq:min1} is equivalent to minimising the following
\begin{align} \label{eq:min3}
  \min_{\mathbf{p}^{(t)}}  \frac{1}{n}\sum_{i=1}^N \frac{1}{p_{i}^t} \big\|\nabla f_i \big(\theta^{(t)}\big) - \nabla f_i(\hat{\theta}) \big\|^2.
\end{align}
The optimal weights which minimise the pseudo-variance are thus given by
\begin{align}  \label{eq:optp2} p_{i}^{t} = \frac{\| \nabla f_i \big(\theta^{(t)}\big) - \nabla f_i(\hat{\theta})) \|}{\sum_{k=1}^N \| \nabla f_k \big(\theta^{(t)}\big) - \nabla f_k(\hat{\theta}) \| } 
\end{align} for $i=1, \ldots, N$. 
\end{lemma}

As in Section~\ref{sec:method_ps}, we can derive a solution to Eq.~\eqref{eq:min3}. However, the resulting sampling algorithm would once again depend on the current state of the chain $\theta^{(t)}$. The process of finding a suitable approximation to the optimal weights given in Eq.~\eqref{eq:optp2} for the control-variate gradient estimator is non-trivial. Our approach will be to choose a set of subsampling weights that could be used for all iterations of the MCMC chain.  

We consider an alternative minimisation problem
 \begin{align} \label{eq:min4}
   \min_{\mathbf{p}^{(t)}, \, p_{i}^t \in [0,1], \sum_i p_{i}^t = 1} \mathbb{E}_\theta \bigg[ \mathbb{V}\big(\tilde{g}^{(t)} \big)\bigg],  \end{align}
where the outer expectation is taken with respect to the posterior distribution. Due to the linearity of expectation, Eq. \eqref{eq:min4} is equivalent to solving the following problem: 
\begin{align*}
 \min_{\mathbf{p}^{(t)}} \mathbb{E}_\theta \bigg[ \frac{1}{n}\sum_{i=1}^N \frac{1}{p_{i}^t} \big\|\nabla f_i \big(\theta^{(t)}\big) - \nabla f_i(\hat{\theta}) \big\|^2 \bigg].
\end{align*} This can easily be shown by using a modified version of the argument given for Lemmas~\ref{lemma3.1} and~\ref{lemma3.2}. The optimal subsampling weights in Eq.~\eqref{eq:optp2} can be approximated by
\begin{align} \label{eq:approxp2}
  p_i \propto \sqrt{\text{tr}\bigg( \nabla^2f_i(\hat{\theta})\,\hat{\Sigma}\,\nabla^2f_i(\hat{\theta})^T\bigg)} \text{ for } i=1,\ldots,N,
\end{align}
where $\nabla^2f_i(\cdot)$ is the Hessian matrix of $f_i(\cdot)$ and $\hat{\Sigma}$ is the covariance matrix of the Gaussian approximation to the target posterior centred at the mode. 

See Appendix \ref{sec:weights_analytical} for a full discussion of how these weights can be obtained analytically and Appendix \ref{sec:app_comp_cost} for an assessment of the computational cost associated with calculating them as a preprocessing step. The computational cost required to calculate the Hessian matrix means that this approach can be computationally expensive for high-dimensional parameters.

\subsection{Adaptive subsampling}
In this section, we present a method for adaptively adjusting the size of the subsample chosen at each iteration. We do this by first finding an upper bound for the pseudo-variance of the stochastic gradient estimator given in Eq.~\eqref{eq:gtilde_cv} and then by rearranging the result to find a lower bound on the subsample size. 

Let us begin by placing a Lipschitz condition on each of the likelihood terms.
\begin{assumption}(Lipschitz continuity of gradients)\\ \label{assumption1}
There exists constants $L_0, \ldots, L_N$ such that 
\begin{align}
    \|\nabla f_i(\theta) - \nabla f_i(\theta') \| \leq L_i \|\theta - \theta'\|
\end{align} for $i=0, \ldots, N$.
\end{assumption}
We can then obtain the following result using Assumption~\ref{assumption1}.
\begin{lemma} \label{lemma3.3} Under Assumptions~\ref{assumption1}, the pseudo-variance of the stochastic gradient estimator  defined in Eq.~\eqref{eq:gtilde_cv} can be bounded above by  
\begin{align} \label{eq:boundcv}  \V(\tilde{g}) \leq \frac{1}{n} \|\theta^{(t)} - \hat{\theta} \|^2 \bigg(\sum_{i=1}^N \frac{L_i^2}{p_i^t}\bigg)  \,. \end{align} where $\mathbf{p}^{(t)} = (p_{1}^t, \ldots, p_{N}^t)^T$ is a set of user-defined discrete weights.
\end{lemma} 

We can minimise the upper bound provided in Eq.~\eqref{eq:boundcv} if we plug in the optimal weights given in Eq.~\eqref{eq:optp2}. In practice, however, it is advantageous to choose the preferential subsampling scheme based on its ease of computation.

The bound provided in Lemma~\ref{lemma3.3} can still be used more generally to control the size of the pseudo-variance of the stochastic gradient estimator given in Eq.~\eqref{eq:gtilde_cv}. If we want to set the upper threshold of the pseudo-variance to be some fixed value $V_0 > 0$, we need to ensure that 
$$\frac{1}{n} \|\theta^{(t)} - \hat{\theta} \|^2 \bigg(\sum_{i=1}^N \frac{L_i^2}{p_i^t}\bigg)  < V_0, \, $$
for all iterations $t = 1, \ldots, T$. We can rearrange the inequality above to obtain the following lower bound on the subsample size, 
\begin{align} \label{eq:nlb} n > \frac{1}{V_0} \|\theta^{(t)} - \hat{\theta} \|^2 \bigg(\sum_{i=1}^N \frac{L_i^2}{p_i^t}\bigg). \end{align}
For a given preferential subsampling scheme $\mathbf{p}^{(t)}$,  we can control the noise of the  stochastic gradient estimator given in Eq.~\eqref{eq:gtilde_cv} by choosing the subsample size $n \propto  \|\theta^{(t)} - \hat{\theta} \|^2$. This means that the subsample size can be set adaptively according to the current state of the chain. 

Our proposed algorithm is provided in Algorithm~\ref{alg:asgldcv}. The subsample size at iteration $t$, $n^{(t)}$, will be updated using the lower bound obtained in Eq.~\eqref{eq:nlb}. For a fixed noise threshold $V_0$, it will be possible to decrease or increase the size of $n^{(t)}$ depending on how far or close $\theta^{(t)}$ is to the posterior mode, $\hat{\theta}$. 

This method is suitable for use on models that satisfy Assumption 1. Appendix \ref{sec:app_models} provides a selection of examples where the Lipschitz constants can be calculated exactly. 

\begin{algorithm}[H]
  \caption{Adaptive SGLD-CV with preferential subsampling (ASGLD-CV-PS)} \label{alg:asgldcv}
  \begin{algorithmic}[1]
    \STATE{Input: initialise $\theta^{(1)}$ close to $\hat{\theta}$, gradients $\nabla f_i \big(\hat{\theta})$, weights $ \mathbf{p}^{(1)}$, step-size  $\epsilon$, noise threshold $V_0$, Lipschitz constants $\{L_i\}_{i=1}^N$.}
   \FOR{$t = 1, 2, \ldots, T$}
   \STATE{Update $\mathbf{p}^{(t)}$.}
   \STATE{Find smallest possible $n^{(t)}$ using Eq.~\eqref{eq:nlb}}.
   \STATE{Sample $n^{(t)}$ indices $\mathcal{S}^t$ according to $\mathbf{p}^{(t)}$ with replacement.}
    \STATE{Calculate $\tilde{g}^{(t)}$ using Eq.~\eqref{eq:gtilde_cv}}
      \STATE{Update parameters according to Eq.~\eqref{eq:sgldupdate}.}
   %\STATE{Update parameters  $\theta^{(t+1)} \leftarrow \theta^{(t)} - \frac{\epsilon}{2} \cdot  \tilde{g}^{(t)} + \N_d(0, \epsilon_t I_{d \times d})$ }
    \ENDFOR 
    \RETURN $\theta^{(T+1)}$ 
 \end{algorithmic}
\end{algorithm}

%%%%%%%%%%%%%%%%%%%%%%%%%%%%%%%%%%%%%%%%%%%%%%%%%%%%%%%%%%%%%%%%%%%%%%%%%%%%%%%%%%%% 

\section{Related work}
\label{sec:relatedwork}
The idea of using a non-uniform discrete distribution to draw subsamples and reweight a gradient estimator has been well-explored within the stochastic optimisation literature. Typically, the aim of these methods is to control the variance of the gradient estimator, in order to improve the speed of convergence of the algorithm. 

Various papers explore the use of a static or time-invariant subsampling schemes.  \cite{zhao2014stochastic, zhao2015stochastic} and \cite{kern2016svrg} propose the use of an \textit{importance sampling} approach for SGD-type algorithms, where the subsampling weights are chosen according to the Lipschitz smoothness constants of $N$ individual cost functions, i.e. $p_i = \frac{L_i}{\sum_{j=1}^N L_j}$. \cite{zhao2014accelerating} consider a \textit{stratified sampling} approach, where data points are assigned the the same weight if they belong to the same strata or cluster. \cite{zhang2017determinantal} propose the use of determinantal point processes to diversify the subsamples selected for SGD, constructing a soft similarity measure to reweight data
points.

Inspired by active learning methods, \cite{salehi2017stochastic} create a multi-armed bandit (MAB)
framework to dynamically update the subsampling weights over several iterations of the SGD algorithm. Feedback is collected via the most recent stochastic gradients and passed into the MAB at each iteration. \cite{liu2020adam} adapts the work of \citeauthor{salehi2017stochastic} and extends it to the minibatch setting for the ADAM algorithm \citep{adam2015}. 

There have only been a handful of papers considering similar ideas within the stochastic gradient MCMC literature. \cite{fu2017cpsg} 
extend the stratified sampling methodology of \cite{zhao2014accelerating} to the general class of SGMCMC algorithms.  \cite{li2021improving} meanwhile propose an  exponentially weighted stochastic gradient method, which can be combined with other variance reduction techniques. 

%%%%%%%%%%%%%%%%%%%%%%%%%%%%%%%%%%%%%%%%%%%%%%%%%%%%%%%%%%%%%%%%%%%%%%%%%%%%%%%%%%%% 

\section{Numerical experiments}
In the experiments to follow, we compare our proposed preferential subsampling approaches in a number of different scenarios. Our aim here is to demonstrate the value of preferential subsampling as a variance control measure. To communicate the idea succinctly, we only include the benchmark methods SGLD and SGLD-CV in our results. 

We evaluate the performance of our proposed methods on both real and synthetic data. Please refer to Appendix \ref{sec:app_models} for detailed information about the datasets considered. 

A fixed step-size scheme for $\epsilon$ is used throughout, as suggested by \cite{vollmer2016non}. To ensure a fair comparison, all samplers were run with the same step-size (with $\epsilon \approx \frac{1}{N}$). This allowed us to control for discretisation error and to independently assess the performance benefits offered by preferential subsampling. See Appendix \ref{sec:app_exp_setup} for further details. 

For samplers where the burn-in phase is replaced by an optimiser, we have opted to use an off-the-shelf implementation of ADAM \citep{adam2015} to find the posterior mode. Unless stated otherwise, all samplers are implemented using sampling with replacement. 

\paragraph{Computing environment} We used the \texttt{jax} autograd module to implement the SGMCMC methods. Our results were obtained on a four-core 3.00GHz Intel Xeon(R) Gold virtual desktop.  The code for this paper is hosted on GitHub\footnote{Repository: \url{https://github.com/srshtiputcha/sgmcmc_preferential_subsampling}}. 

\label{sec:experiments}
\subsection{Models}
We compare sampler performance on the following three models: (i) bivariate Gaussian, (ii) binary logistic regression and (iii) linear regression. Full model details (including the derivation gradients and Lipschitz constants) are provided in Appendix \ref{sec:app_models}.

The examples have been deliberately chosen to be simple and our reasons for doing so are threefold. Firstly, our proposed methods rely upon being able to estimate the posterior mode well and as such, we are prioritising
models where the mode is easy to find. Secondly, the Lipschitz constants for these models are known and this allows us to test our adaptive subsampling approach. And lastly, the SGLD-CV-PS
subsampling scheme outlined in Eq.~\eqref{eq:approxp2} requires the Hessian matrix, $\nabla^2 f_i(\cdot)$, to be computed for all data points and this is a costly preprocessing step for large parameter spaces\footnote{We have provided an extended discussion of the preprocessing costs associated with SGLD-CV-PS in Appendix \ref{sec:app_comp_cost}.}. 
\subsubsection{Bivariate Gaussian}
We simulate independent data from
$X_i|\theta \sim \mathcal{N}_2(\theta, \Sigma_x)$ for $i=1,\ldots,N.$ 
It is assumed that $\theta$ is unknown and $\Sigma_x$ is known. 
The conjugate prior for $\theta$ is set to be $\theta \sim \mathcal{N}_2(\mu_0, \Lambda_0). $ The prior hyperparameters of the prior are $\mu_0 = (0,0)^T$and $\Lambda_0 = \mbox{diag}(1\times10^3,2)$. The target posterior is a non-isotropic Gaussian with negatively correlated parameters. 
\subsubsection{Binary logistic regression}
Suppose we have data $x_1, \dots, x_N$ of dimension $d$ taking values in $\mathbb R^{d}$, where each $x_i = (1, x_{i1}, \ldots, x_{ip})^T$ (where $d=p+1$). Let us suppose that we also have the corresponding response variables $y_1, \dots, y_N$ taking values in $\{0, 1\}$. Then a logistic regression model with parameters $\theta=(\beta_0,\beta_1, \ldots, \beta_{p})$ will have the following density function
$$  p(y_i | x_i, \theta ) = \bigg( \frac{1}{1 + e^{-\theta^T x_i}}\bigg)^{y_i} \bigg( 1 - \frac{1}{1 + e^{-\theta^T x_i}} \bigg)^{1-y_i}.$$

The prior for $\theta$ is set to be $\theta \sim \mathcal{N}_d(\mu_0, \Lambda_0).$ The hyperparameters of the prior are $\mu_0 = (0, \ldots,0)^T$ and $\Lambda_0 = \mbox{diag}(10,d)$. 

\subsubsection{Linear regression}
Suppose we have data $x_1, \dots, x_N$ of dimension $d$ taking values in $\mathbb R^{d}$, where each $x_i = (1, x_{i1}, \ldots, x_{ip})^T$ ($d=p+1$). Let us suppose that we also have the corresponding response variables $y_1, \dots, y_N$ taking values on the real line. 

We define the following linear regression model,
$$ y_i = x_i^T \theta + \eta_i, \enspace \eta_i \sim \mathcal{N}(0, 1),$$
with parameters $\theta=(\beta_0,\beta_1, \ldots, \beta_{p})$. The prior for $\theta$ is the same as above. 
\subsection{Metrics}
We assess the performance of our samplers using the following metrics. 
\subsubsection{Kullbeck-Leibler (KL) divergence}
The KL divergence is a measure of difference between two probability distributions with densities $p(\cdot)$ and $q(\cdot)$ and is given by, 
$$D_{KL}(p || q) = \int p(\theta) \log \frac{p(\theta)}{q(\theta)} d\theta.$$
In the case of our bivariate Gaussian model, we know that the target posterior is conjugate and the KL divergence between two Gaussians can be written analytically. We use the KL divergence to measure the difference between the target posterior and our generated samples in Figure~\ref{fig:sgld_comp}(a).
\subsubsection{Log-loss}
The log-loss is a popular metric for assessing the predictive accuracy of the logistic regression model on a test dataset, $\mathcal{T}^*$. For binary classification, the log-loss is given by
$$l(\theta, \mathcal{T}^*) = -\frac{1}{|\mathcal{T}^*|} \sum_{i \in |\mathcal{T}^*|} \log p(y_i^* | x_i^*, \theta). $$
We compute the log-loss for our logistic regression example in  Figure~\ref{fig:sgldcv_comp}(b)(ii).
\subsubsection{Kernel Stein discrepancy}
We measure the sample quality of our MCMC chains using the kernel Stein discrepancy (KSD). The KSD assesses the discrepancy between the target posterior $\pi$ and the empirical distribution $\tilde{\pi}_K$ formed by SGMCMC samples $\{\theta\}_{k=1}^K$  \citep{liu2016kernelized, gorham2017measuring}. A key benefit of the KSD is that it penalises the bias present in our MCMC chains. We can define the KSD as,
\begin{align}
  KSD(\tilde{\pi}_K, \pi) = \sum_{j=1}^d \sqrt{\sum_{k, k' = 1}^K \frac{k_j^0(\theta_k, \theta_{k'})}{K^2}},
\end{align} where the Stein kernel for $j \in \{1, \ldots, d\}$ is given by,
\begin{align} \label{eq:k0}
  k_j^0(\theta, \theta') = \frac{1}{\pi(\theta)\pi(\theta')} \nabla_{\theta_j} \nabla_{\theta'_j} \big(\pi(\theta)\mathcal{K}(\theta, \theta')\pi(\theta')\big)
\end{align} and $\mathcal{K}(\cdot, \cdot)$ is a valid kernel function.~\citet{gorham2017measuring} recommend using the inverse multi-quadratic kernel, $\mathcal{K}(\theta, \theta') = \big(c^2 + \| \theta - \theta' \|_2^2\big)^\beta$, which detects non-convergence for $c >0 $ and $\beta \in (-1,0)$. In practice, the full-data gradients in Eq.~\eqref{eq:k0} can be replaced by noisy, unbiased estimates. We compute the KSD for our linear and logistic regression examples in Figures~\ref{fig:sgldcv_comp}(a) - (b)(i),~\ref{fig:asgldcv_synth}(a),~\ref{fig:sgld_comp}(b) and~\ref{fig:asgldcv_casp}(a).

\subsection{Numerical results}

\subsubsection{Evaluating the quality of the stochastic gradients}
\label{sec:exp_1}

%-----Figure1-------------%
\begin{figure*}[ht]
\begin{center}
\vspace{-14pt}
\begin{minipage}[c]{.32\textwidth}
\centering 
\begin{minipage}[c]{\textwidth}
  \centering
\includegraphics[width=\textwidth]{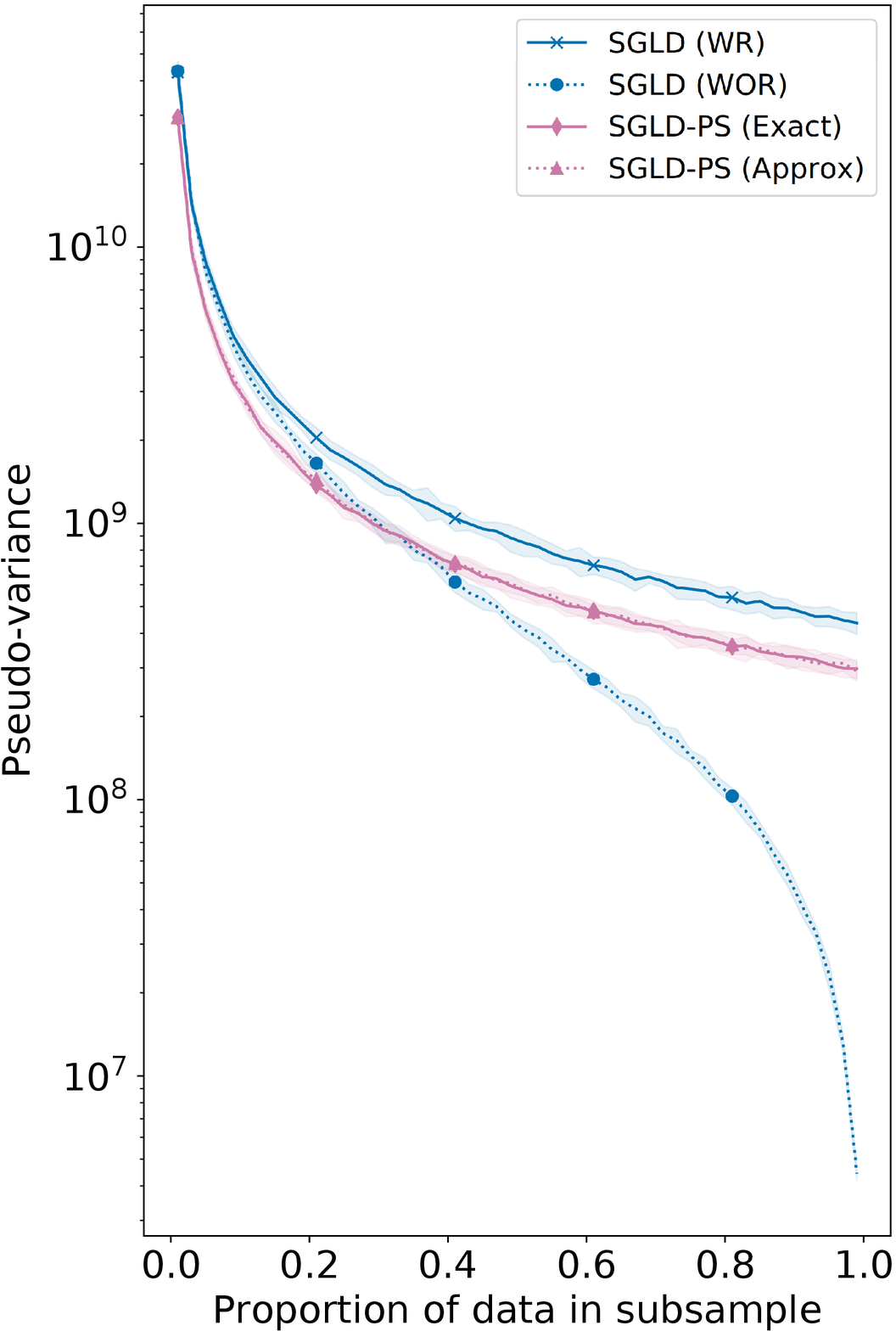}
\vspace{-10pt}
\end{minipage}
(a) 
\end{minipage} 
\begin{minipage}[c]{.32\textwidth}
\centering 
\begin{minipage}[c]{\textwidth}
  \centering
\includegraphics[width=\textwidth]{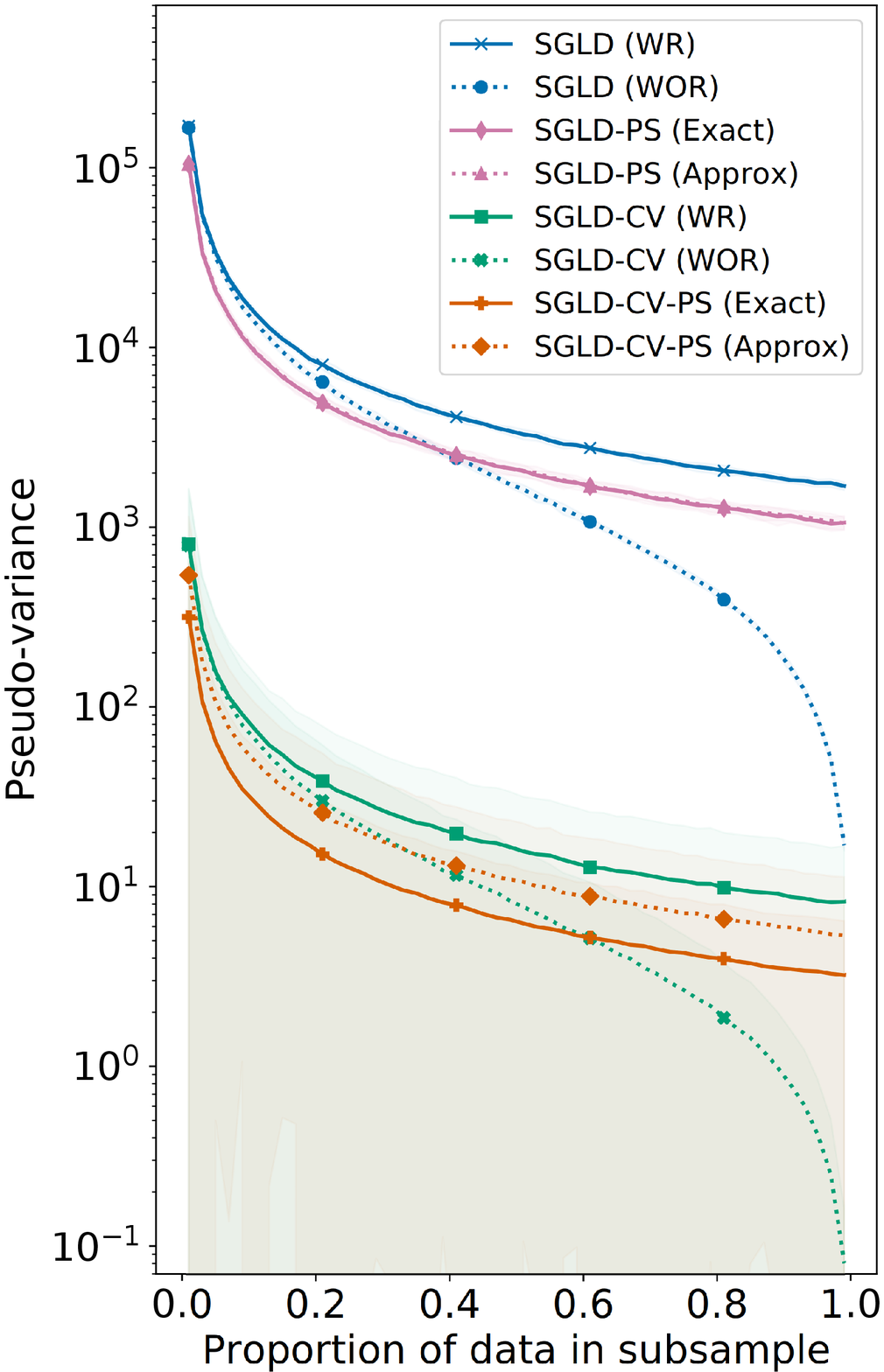}
\vspace{-10pt}
\end{minipage}
(b) 
\end{minipage} 
\begin{minipage}[c]{.32\textwidth}
\centering 
\begin{minipage}[c]{\textwidth}
  \centering
\includegraphics[width=\textwidth]{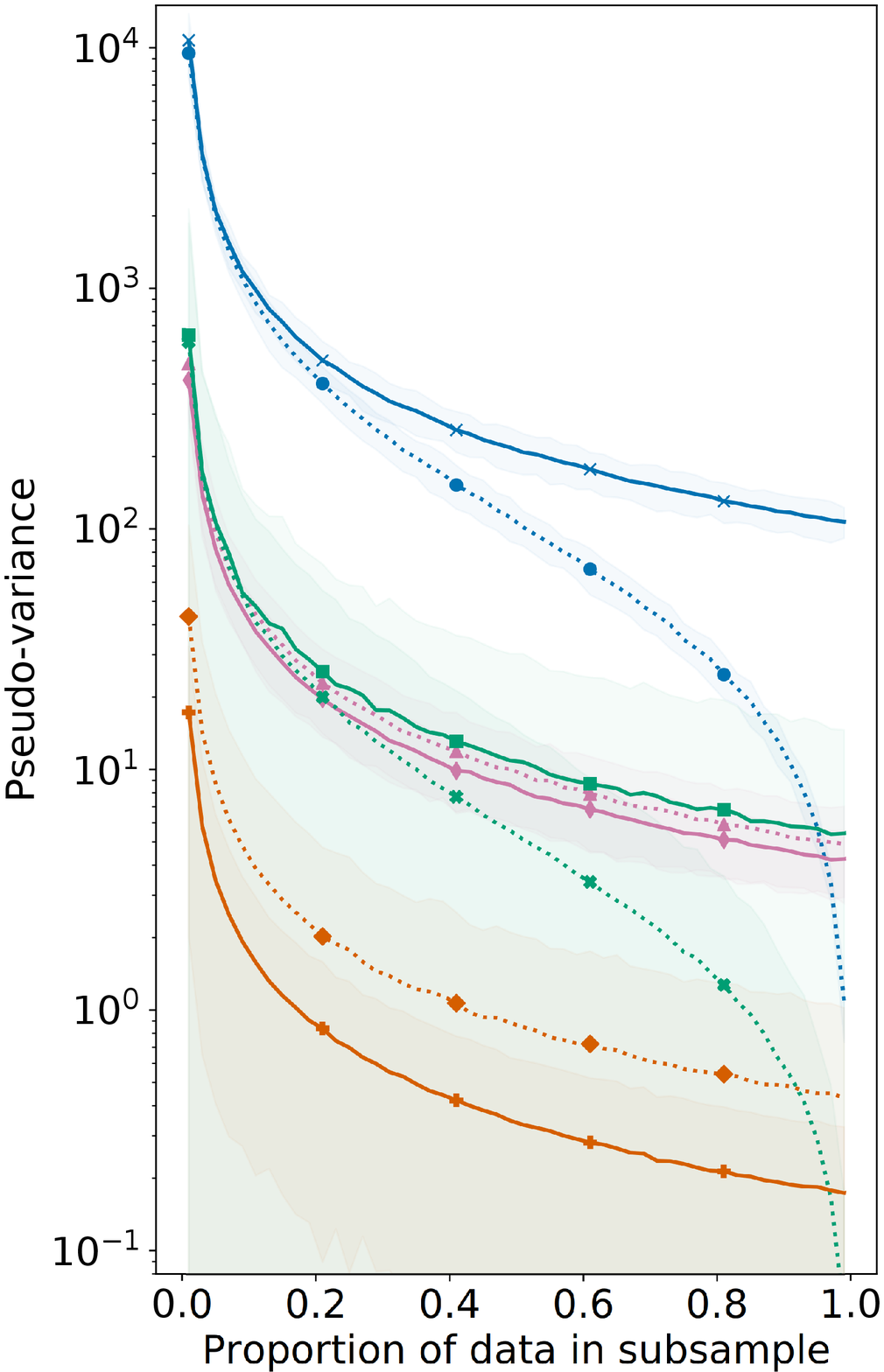}
\vspace{-10pt}
\end{minipage}
(c)
\end{minipage}
\caption{Empirical pseudo-variance against proportion of data in a subsample, $\frac{n}{N}$. (a) bivariate Gaussian, (b) balanced logistic regression, (c) imbalanced logistic regression}
\label{fig:grad_comp}
\end{center} \vspace{-6pt}
\end{figure*}
In this experiment, our objective was to compare the pseudo-variance of our proposed gradient estimators against the proportion of data used in a subsample, $\frac{n}{N}$. We consider three scenarios: (a) bivariate Gaussian (b) balanced bivariate logistic regression, and (c) imbalanced bivariate logistic regression. For ease of computation, a synthetic dataset of size $N=10^3$ was used for all models.

In each scenario, we generated ten candidate draws of $\theta$ and calculated an empirical estimate of the pseudo-variance at each. The candidate draws were sampled either from the posterior (for scenario (a)) or from a normal approximation to the posterior (for scenarios (b) and (c)). We plot the mean empirical estimate of the pseudo-variance for various subsample sizes. 

Figure~\ref{fig:grad_comp}(a) compares the stochastic gradients for SGLD with and without replacement (WR and WOR respectively) and for SGLD-PS with exact and approximate subsampling schemes. Figures~\ref{fig:grad_comp}(b) and (c) additionally compare the gradient estimators of SGLD-CV and those of SGLD-CV-PS with exact and approximate subsampling schemes.\footnote{In the case of a  Gaussian posterior, the SGLD-CV stochastic gradient offers optimal variance reduction, with optimal weights $p_i = \frac{1}{N}
$. For this reason, there is no extra improvement gain to be obtained here by implementing SGLD-CV-PS}

SGLD and SGLD-CV with replacement offer the best variance reduction for larger subsamples, as $\frac{n}{N}$ tends towards 1. For more reasonable subsample sizes of $n \leq 0.2N$, however, there is no major benefit in generating subsamples without replacement. Figure~\ref{fig:grad_comp} illustrates that there is a marked reduction in the pseudo-variance when a preferential subsampling scheme is used.

SGLD-PS and SGLD-CV-PS consistently outperform their vanilla counterparts and there seems to be very little difference between the exact and approximate schemes for SGLD-PS. Whereas, there is a difference in performance between the exact and approximate subsampling schemes for SGLD-CV-PS. This difference is noticeable in Figure~\ref{fig:grad_comp}(c) for the synthetic imbalanced logistic regression data. Practically, it is not feasible to use the exact subsampling weights, but as illustrated here, using approximate preferential weights is always better than using uniform weights.

\subsubsection{Performance of fixed subsample size methods} \label{sec:exp_2}
\begin{figure*}[h]
\begin{center}
\vspace{-5pt}
\begin{minipage}[c]{.31\textwidth}
\centering 
\begin{minipage}[c]{\textwidth}
  \centering
\includegraphics[width=\textwidth]{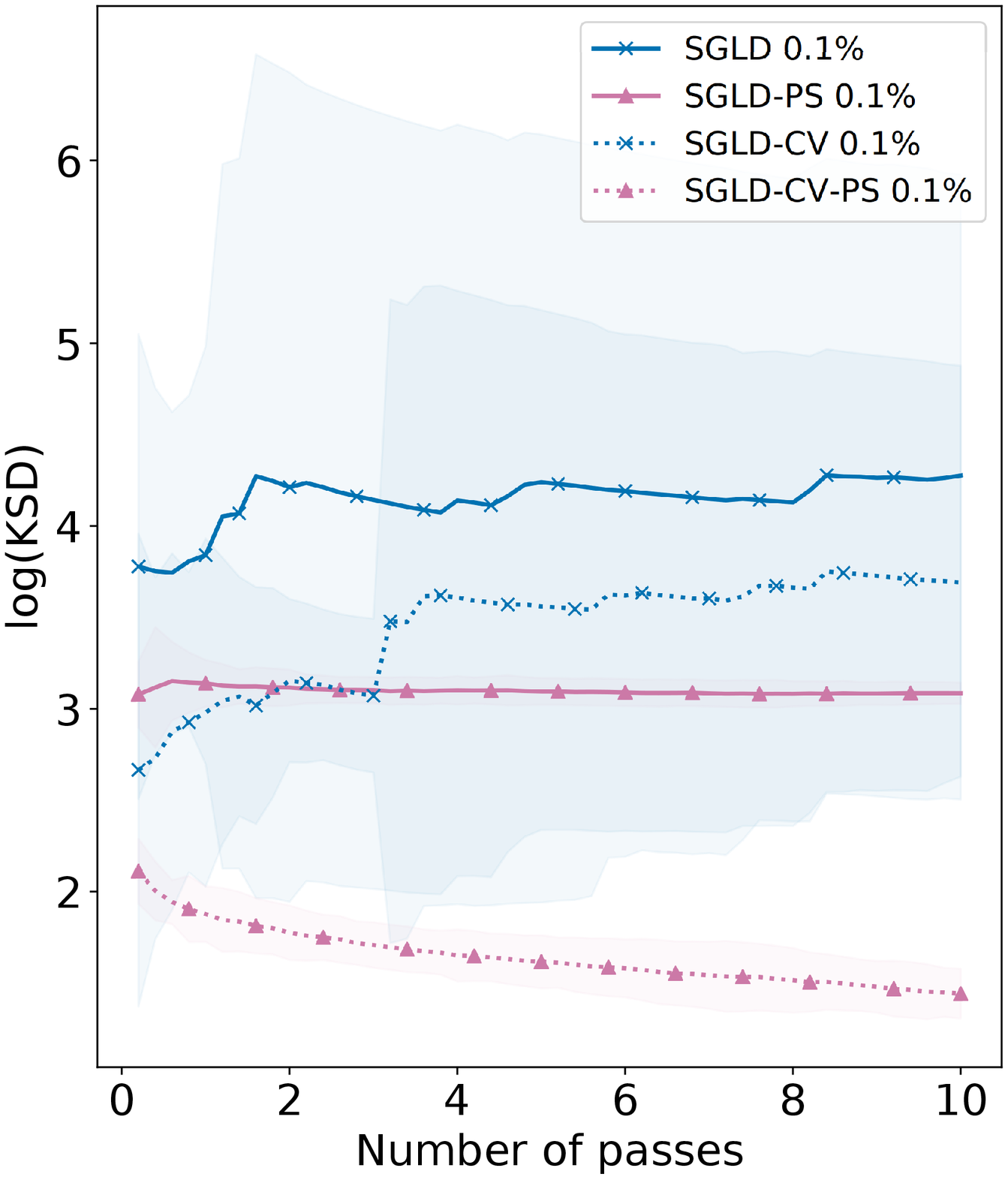} 
\vspace{-5pt}
\end{minipage}
(a) 
\end{minipage} 
\begin{minipage}[c]{.32\textwidth}
\centering 
\begin{minipage}[c]{\textwidth}
  \centering
\includegraphics[width=\textwidth]{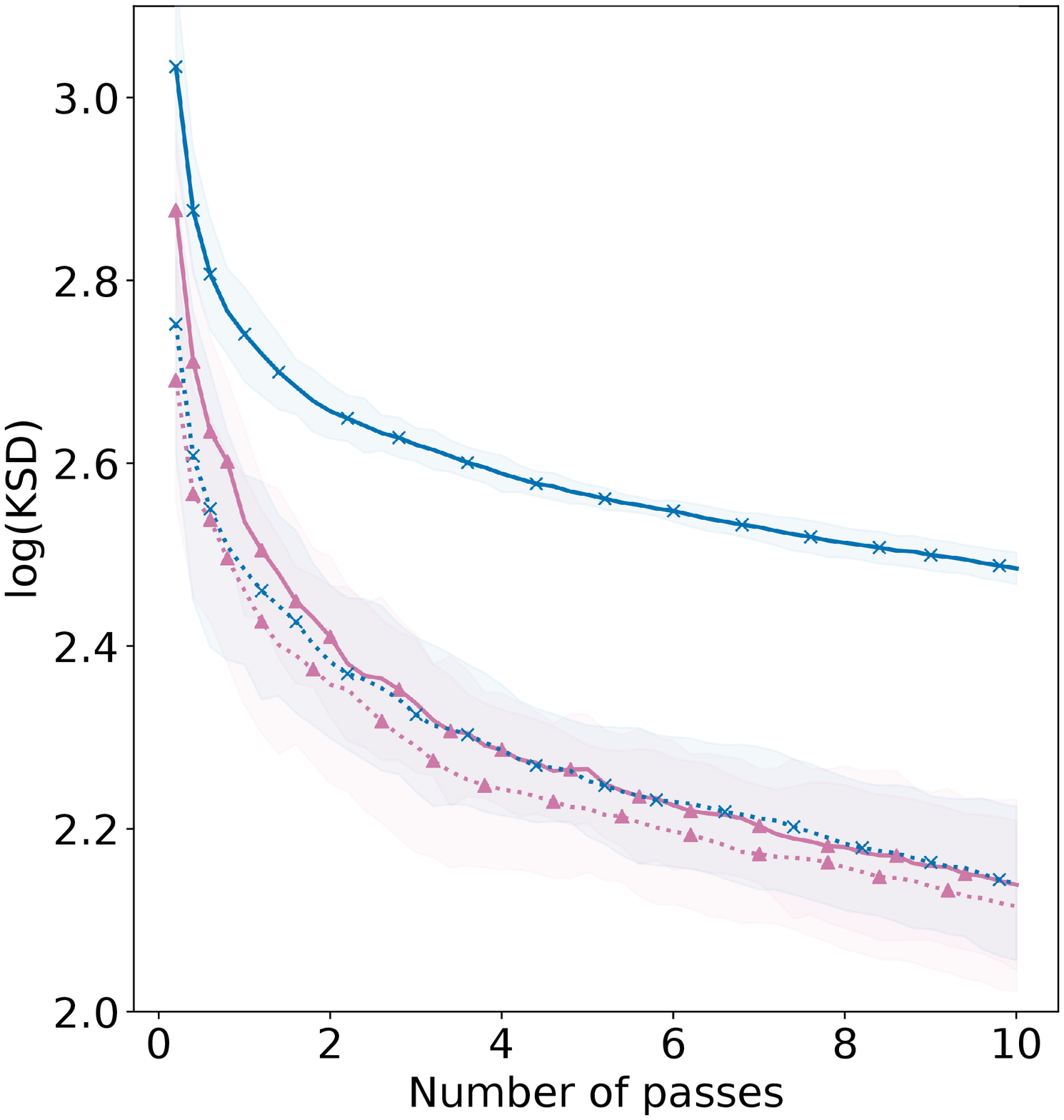} 
\vspace{-10pt}
\end{minipage} 
(b)(i)
\end{minipage} 
\begin{minipage}[c]{.32\textwidth}
\centering 
\begin{minipage}[c]{\textwidth}
  \centering
\includegraphics[width=\textwidth]{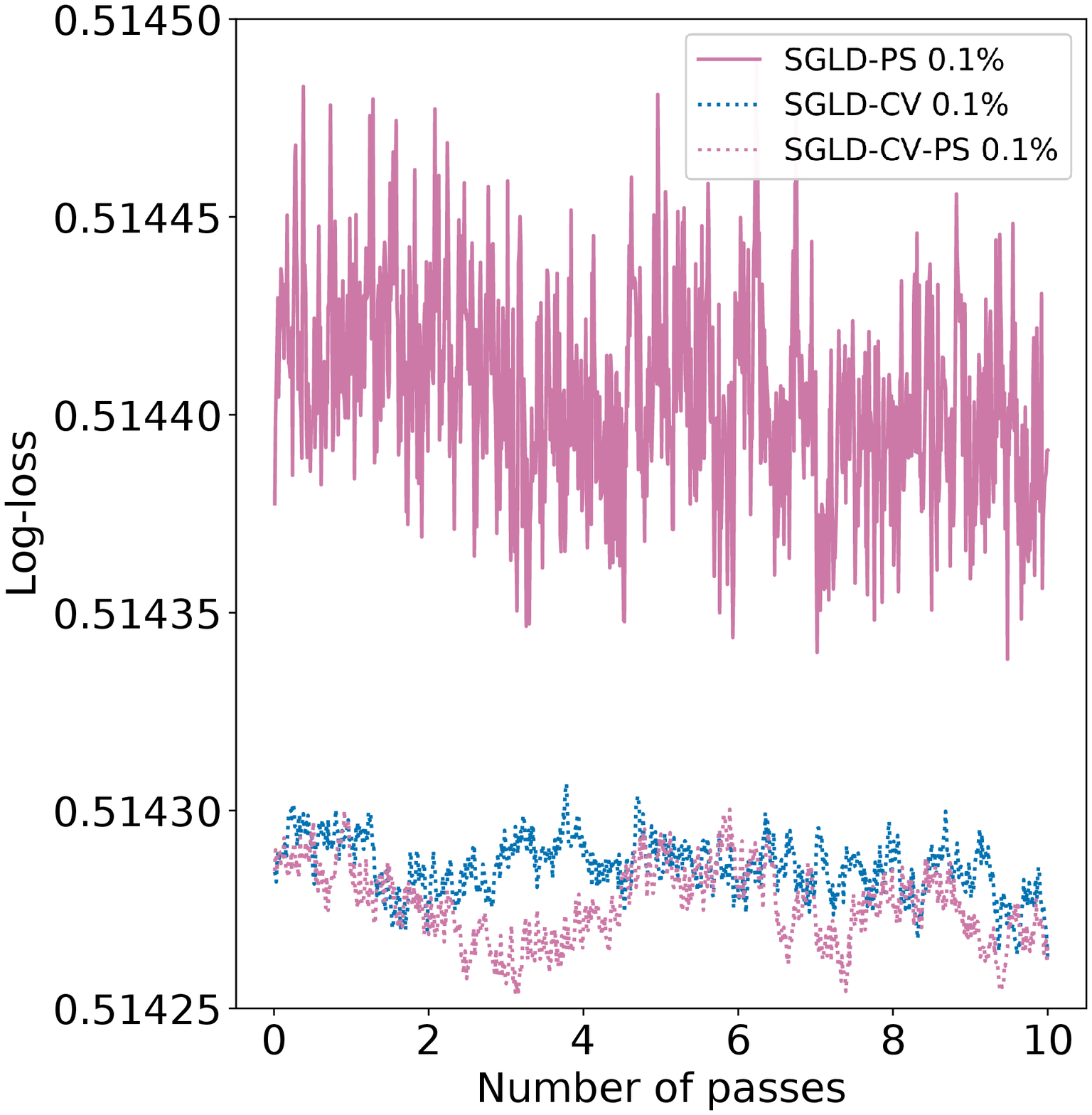} 
\vspace{-10pt}
\end{minipage}
(b)(ii)
\end{minipage} 
\caption{Sampler performance of SGLD, SGLD-CV, SGLD-PS and SGLD-CV-PS for 0.1\% subsample size over 10 passes through the data. (a) linear regression model on the CASP data (y-axis: KSD); (b) logistic regression on the covertype data (y-axis: (i) KSD, (ii) log-loss).}
\label{fig:sgldcv_comp}
\end{center} \vspace{-6pt}
\end{figure*}

In Figure~\ref{fig:sgldcv_comp}, we compare the sampler performance of SGLD, SGLD-CV, SGLD-PS and SGLD-CV-PS for a subsample size of 0.1\% of the dataset size over 10 passes of the data. We have run ten MCMC chains allowing for an equal number of iterations for both burn-in and sampling. 

Figure~\ref{fig:sgldcv_comp}(a) plots the KSD results for the linear regression model fitted on the CASP dataset. The CASP dataset has been obtained from the UCI Machine Learning repository and contains 45,730 instances and 9 features. 

Furthermore, Figure~\ref{fig:sgldcv_comp}(b) plots the results of fitting the logistic regression model to the covertype dataset \citep{blackard1998}. The covertype dataset contains 581,012 instances and 54 features. The KSD results are shown in Figure~\ref{fig:sgldcv_comp}(b)(i) and the log-loss (evaluated every 10 iterations) is computed on the test set in Figure~\ref{fig:sgldcv_comp}(b)(ii). 

In addition, we separately compare the sampler performance of SGLD and SGLD-PS in Figure~\ref{fig:sgld_comp} (see Appendix \ref{sec:app_add_exp}) for subsample sizes of $1\%$, $5\%$ and $10\%$ of the dataset size, over 500 passes of the data. As before, ten MCMC chains were run for each subsample size tested, allowing for an equal number of iterations for both burn-in and sampling 

Figure~\ref{fig:sgld_comp}(a) plots the KL divergence for the bivariate Gaussian model fitted on synthetic data of size $N=10^4$. Figure~\ref{fig:sgld_comp}(b) plots the KSD for the logistic regression model fitted on the covertype dataset \citep{blackard1998}.

We can see that SGLD-CV-PS exhibits the best performance overall. More generally, we find that there is a benefit in implementing preferential subsampling for vanilla SGLD as well. In practice, we find that the largest performance gains are found for preferential subsampling when the subsampling size is reasonably small.

%-----Figure2-------------%

\subsubsection{Performance of adaptive subsampling} 
\label{sec:exp_3}

We are interested in assessing the sampler performance of ASGLD-CV and ASGLD-CV-PS over $10^4$ iterations. In this experiment, we considered two scenarios: (i) the logistic regression model on balanced synthetic data of size $N = 10^4$; and (ii) the linear regression model on the CASP data. The results for scenario (i) are plotted in Figure~\ref{fig:asgldcv_synth}. Please refer to Figure~\ref{fig:asgldcv_casp} in Appendix \ref{sec:app_add_exp} for the results of scenario (ii). Both models satisfy Assumption~\ref{assumption1}. 

In order to implement the adaptive subsampling methods, we had to first pick a suitable pseudo-variance threshold, $V_0$. We generated ten chains of SGLD-CV and SGLD-CV-PS for a fixed subsample of size 0.1\% of the dataset size. For each chain, we then: \begin{enumerate}
    \item calculated the squared Euclidean distances between the mode and the samples,  $\| \theta - \hat{\theta}\|^2$,
    \item found the 95-th percentile of the array of squared distances, and \item calculated a proposal for $V_0$ using the bound in Eq.~\eqref{eq:boundcv}.
\end{enumerate} We set $V_0$ to be the largest proposal amongst the ten chains. 

Figure~\ref{fig:asgldcv_synth}(a) plots the KSD for all four methods and Figure~\ref{fig:asgldcv_synth}(b) displays the adaptive subsample sizes selected along one chain of ASGLD-CV-PS. Figure~\ref{fig:asgldcv_synth}(c) compares the number of passes through the data considered by fixed subsampling versus ASGLD-CV-PS over $10^4$ iterations.  

Overall, we see that the performance of ASGLD-CV-PS is somewhat better than that of ASGLD-CV. We cannot always presume that the adaptive subsampling methods will outperform their fixed subsampling counterparts (see Figure~\ref{fig:asgldcv_casp}(a) for instance) in terms of sample quality. However, it is clear that the adaptive subsampling methods successfully process far less of the data over $10^4$ iterations with no significant reduction in statistical accuracy.

%-----Figure3-------------%
\begin{figure*}[h]
\begin{center}
\vspace{-5pt}
\begin{minipage}[c]{.32\textwidth}
\centering 
\begin{minipage}[c]{\textwidth}
  \centering
\includegraphics[width=\textwidth]{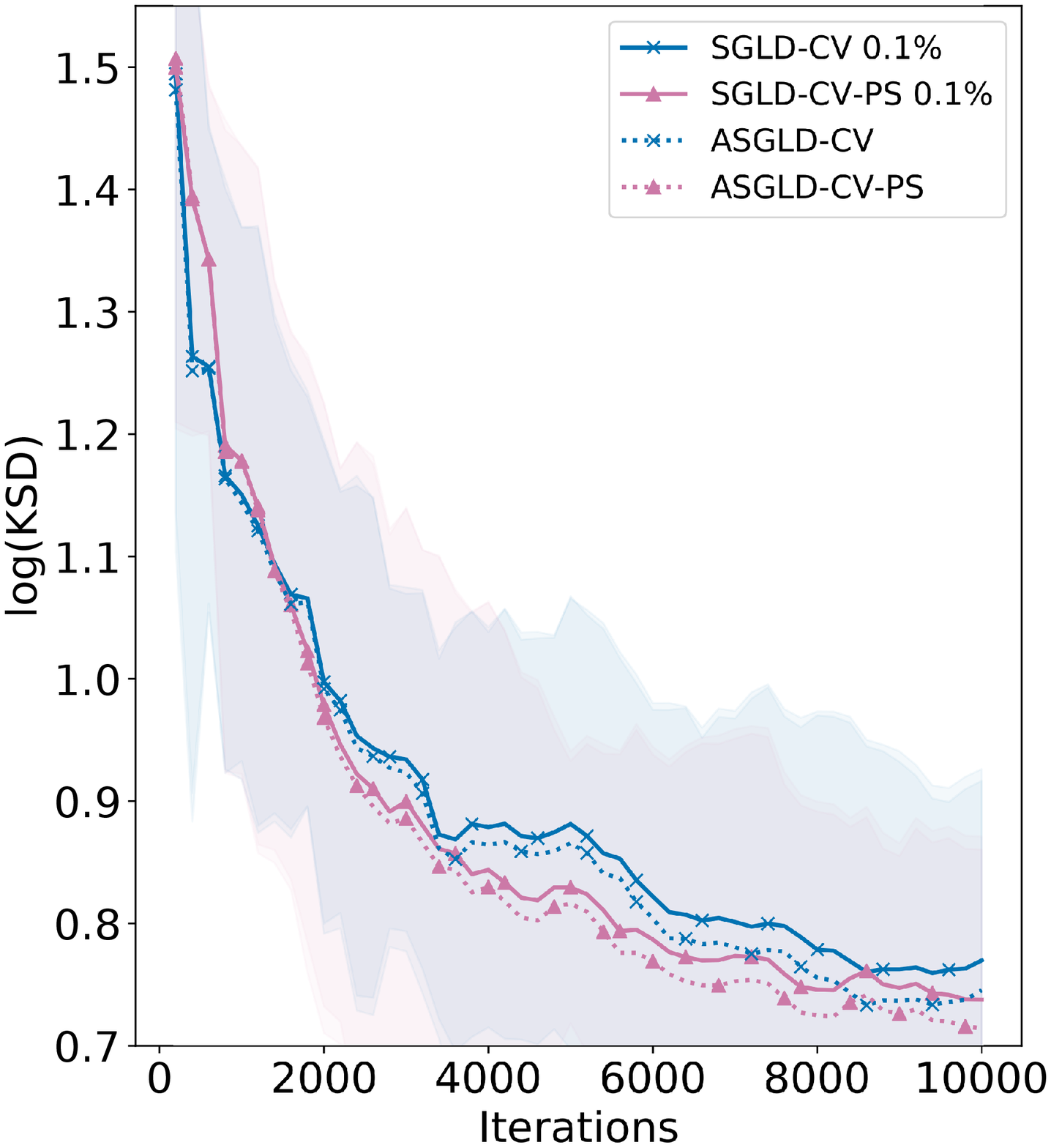} 
\vspace{-10pt}
\end{minipage}
(a) 
\end{minipage} 
\begin{minipage}[c]{.32\textwidth}
\centering 
\begin{minipage}[c]{\textwidth}
  \centering
\includegraphics[width=\textwidth]{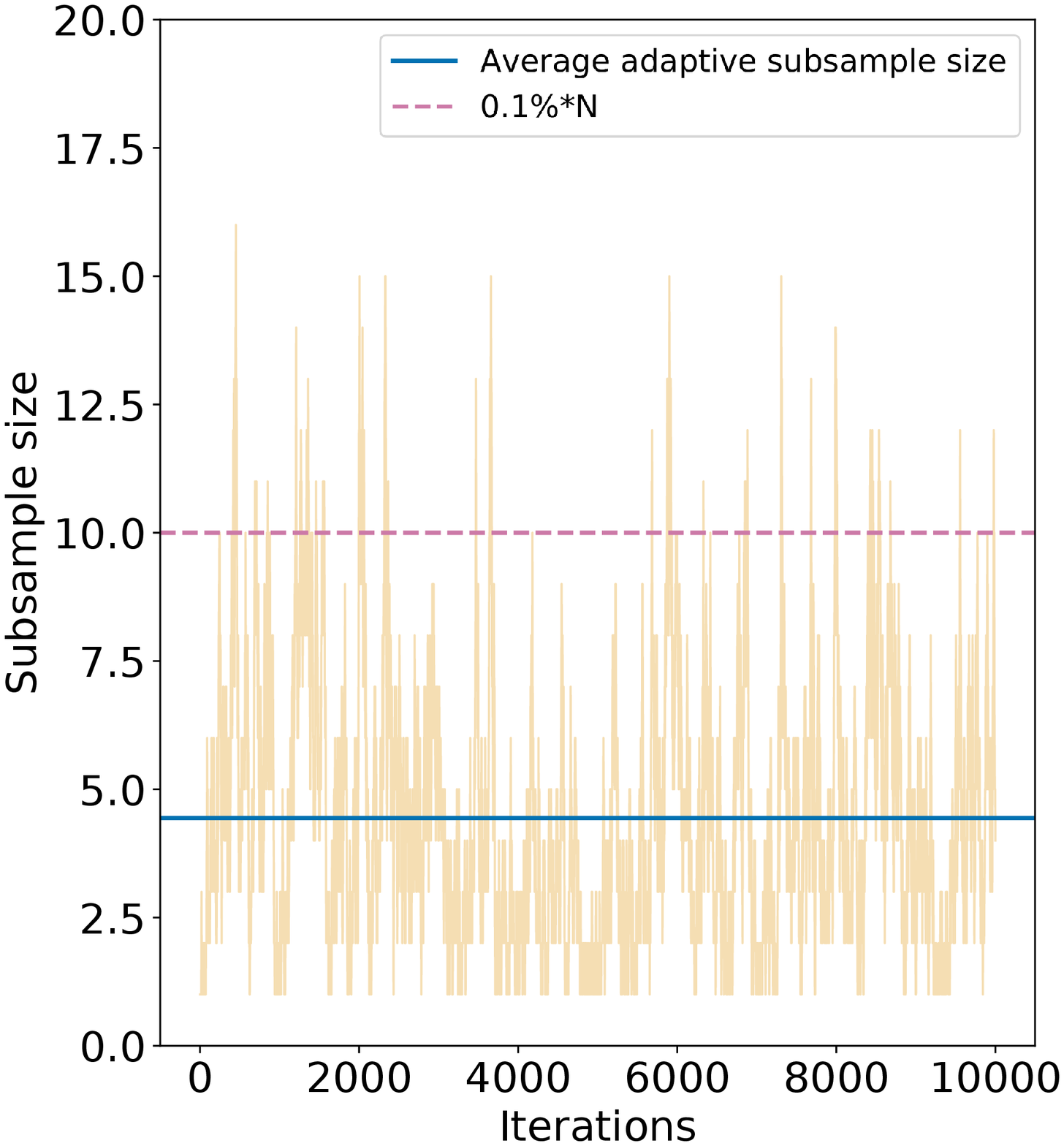}
\vspace{-10pt}
\end{minipage}
(b)
\end{minipage}
\begin{minipage}[c]{.32
\textwidth}
\centering 
\begin{minipage}[c]{\textwidth}
  \centering
\includegraphics[width=\textwidth]{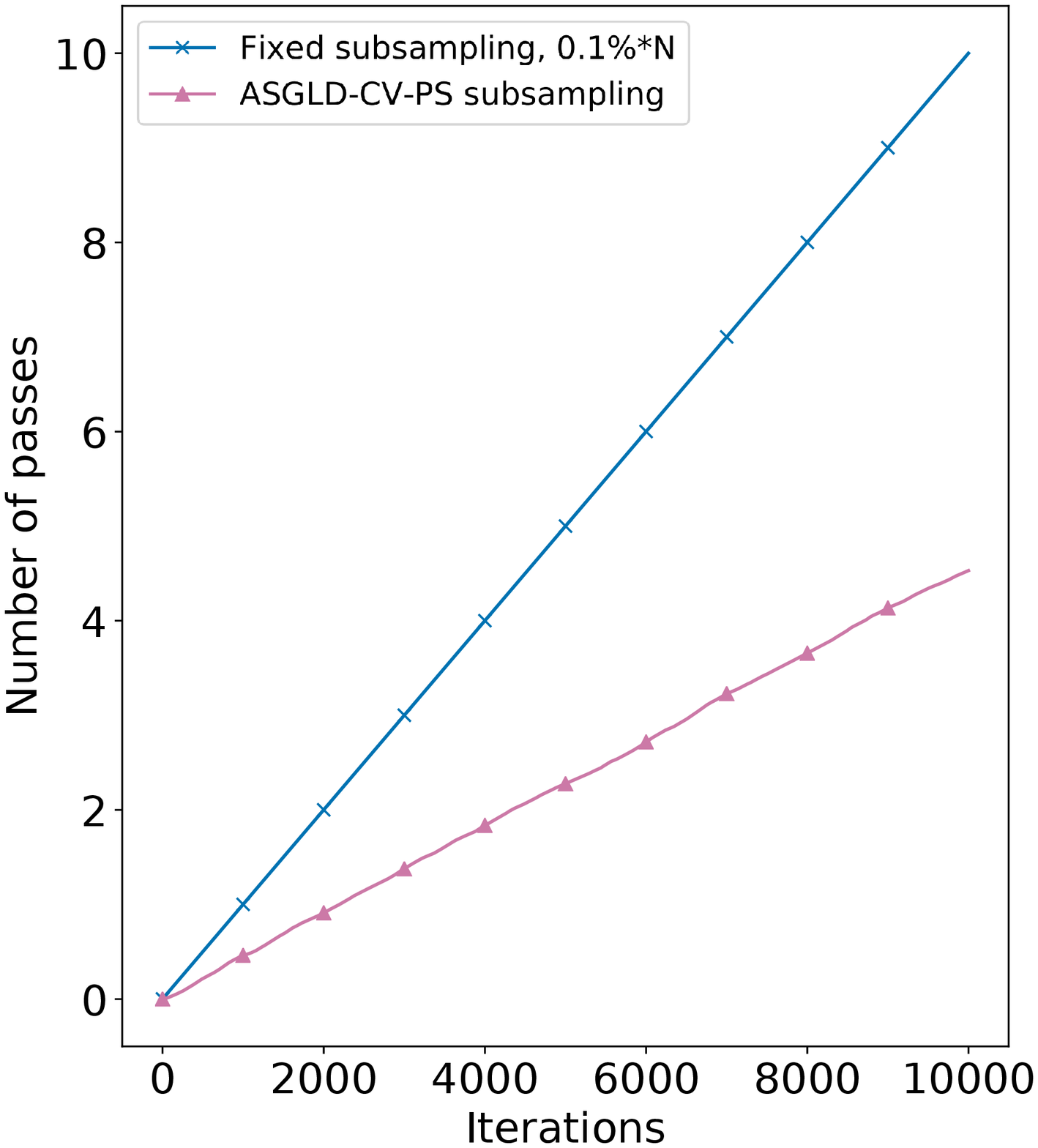} 
\vspace{-10pt}
\end{minipage}
(c)
\end{minipage}
\caption{A logistic regression model fitted on balanced synthetic data of size $N=10^4$. (a) KSD comparison of SGLD-CV, SGLD-CV-PS, ASGLD-CV and ASGLD-CV-PS over $10^4$ iterations; (b) adaptive subsample sizes selected along one ASGLD-CV-PS chain; (c) the number of passes through the data considered by fixed versus adaptive subsampling.}
\label{fig:asgldcv_synth}
\end{center}
\end{figure*} 

%%%%%%%%%%%%%%%%%%%%%%%%%%%%%%%%%%%%%%%%%%%%%%%%%%
\section{Conclusion}
%%%%%%%%%%%%%%%%%%%%%%%%%%%%%%%%%%%%%%%%%%%%%%%%%
We have used preferential subsampling to reduce the variance of the stochastic gradient estimator for both SGLD and SGLD-CV. In addition, we have extended SGLD-CV to allow for adaptive subsampling.

We have empirically studied the impact of preferential subsampling on a range of synthetic and real-world datasets. Our numerical experiments successfully demonstrate the performance improvement from both preferential subsampling and adaptively selecting the subsample size. 

Future work in this area could explore the potential for using multi-armed bandits to preferentially select data subsamples for SGMCMC. These concepts have only been previously considered within the context of stochastic optimisation \citep{salehi2017stochastic,liu2020adam}.

The methods outlined in this paper are not limited to Langevin dynamics and can be applied to other SGMCMC samplers. 
Furthermore, it would be worth considering how preferential subsampling could be extended for other variance control methods, such as SAGA-LD or SVRG-LD \citep{dubey2016variance}. This could potentially be done by adapting the ideas presented in \cite{schmidt15}, \cite{kern2016svrg} and \cite{schmidt2017minimizing}. Arguments that mirror those presented in Lemmas~\ref{lemma3.1} and~\ref{lemma3.2} could be used to obtain the optimal preferential subsampling scheme in each new case.

\subsubsection*{Acknowledgements}
SP gratefully acknowledges the support of the EPSRC funded EP/L015692/1 STOR-i Centre for Doctoral Training. CN acknowledges the support of EPSRC grants EP/S00159X/1, EP/V022636/1 and EP/R01860X/1. PF acknowledges support of EPSRC grants EP/N031938/1 and EP/R018561/1.SP gratefully acknowledges the support of the EPSRC funded EP/L015692/1 STOR-i Centre for Doctoral Training. CN acknowledges the support of EPSRC grants EP/V022636/1 and EP/R01860X/1. PF acknowledges support of EPSRC grants EP/N031938/1 and EP/R018561/1.

%Bibliography
\bibliographystyle{plainnat}
\bibliography{references} 

\begin{thebibliography}{35}
\providecommand{\natexlab}[1]{#1}
\providecommand{\url}[1]{\texttt{#1}}
\expandafter\ifx\csname urlstyle\endcsname\relax
  \providecommand{\doi}[1]{doi: #1}\else
  \providecommand{\doi}{doi: \begingroup \urlstyle{rm}\Url}\fi

\bibitem[Baker et~al.(2019)Baker, Fearnhead, Fox, and Nemeth]{baker2019control}
Jack Baker, Paul Fearnhead, Emily~B Fox, and Christopher Nemeth.
\newblock Control variates for stochastic gradient {MCMC}.
\newblock \emph{Statistics and Computing}, 29\penalty0 (3):\penalty0 599--615,
  2019.

\bibitem[Blackard and Dean(1998)]{blackard1998}
Jock~A Blackard and Denis~J Dean.
\newblock Comparative {A}ccuracies of {N}eural {N}etworks and {D}iscriminant
  {A}nalysis in {P}redicting {F}orest {C}over {T}ypes from {C}artographic
  {V}ariables.
\newblock In \emph{Second Southern Forestry GIS Conference}, pages 189--199,
  1998.

\bibitem[Bottou et~al.(2018)Bottou, Curtis, and Nocedal]{bottou2018}
L{\'e}on Bottou, Frank~E Curtis, and Jorge Nocedal.
\newblock Optimization {M}ethods for {L}arge-{S}cale {M}achine {L}earning.
\newblock \emph{SIAM Review}, 60\penalty0 (2):\penalty0 223--311, 2018.

\bibitem[Chatterji et~al.(2018)Chatterji, Flammarion, Ma, Bartlett, and
  Jordan]{chatterji2018theory}
Niladri~S Chatterji, Nicolas Flammarion, Yi-An Ma, Peter~L Bartlett, and
  Michael~I Jordan.
\newblock On the {T}heory of {V}ariance {R}eduction for {S}tochastic {G}radient
  {M}onte {C}arlo.
\newblock In \emph{Proceedings of the 35th International Conference on Machine
  Learning}, volume~80 of \emph{Proceedings of Machine Learning Research},
  pages 764--773. PMLR, 10--15 Jul 2018.

\bibitem[Chen et~al.(2019)Chen, Wang, Zhang, Su, and
  Carin]{chen2019convergence}
Changyou Chen, Wenlin Wang, Yizhe Zhang, Qinliang Su, and Lawrence Carin.
\newblock A {C}onvergence {A}nalysis for {A} {C}lass of {P}ractical
  {V}ariance-{R}eduction {S}tochastic {G}radient {MCMC}.
\newblock \emph{Science China Information Sciences}, 62\penalty0 (12101), 2019.

\bibitem[Chen et~al.(2014)Chen, Fox, and Guestrin]{chen2014}
Tianqi Chen, Emily Fox, and Carlos Guestrin.
\newblock Stochastic {G}radient {H}amiltonian {M}onte {C}arlo.
\newblock In \emph{Proceedings of the 31st International Conference on Machine
  Learning}, volume~32 of \emph{Proceedings of Machine Learning Research},
  pages 1683--1691. PMLR, 2014.

\bibitem[Dalalyan and Karagulyan(2019)]{dalalyan2019}
Arnak~S. Dalalyan and Avetik Karagulyan.
\newblock User-friendly guarantees for the {L}angevin {M}onte {C}arlo with
  inaccurate gradient.
\newblock \emph{Stochastic Processes and their Applications}, 129\penalty0
  (12):\penalty0 5278--5311, 2019.

\bibitem[Dubey et~al.(2016)Dubey, Reddi, Williamson, Poczos, Smola, and
  Xing]{dubey2016variance}
Kumar~Avinava Dubey, Sashank~J Reddi, Sinead~A Williamson, Barnabas Poczos,
  Alexander~J Smola, and Eric~P Xing.
\newblock Variance {R}eduction in {S}tochastic {G}radient {L}angevin
  {D}ynamics.
\newblock In \emph{Advances in Neural Information Processing Systems},
  volume~29, 2016.

\bibitem[Durmus and Moulines(2019)]{durmus2019high}
Alain Durmus and Eric Moulines.
\newblock High-dimensional {B}ayesian inference via the unadjusted {L}angevin
  algorithm.
\newblock \emph{Bernoulli}, 25\penalty0 (4A):\penalty0 2854--2882, 2019.

\bibitem[Dwivedi et~al.(2018)Dwivedi, Chen, Wainwright, and Yu]{dwivedi2018log}
Raaz Dwivedi, Yuansi Chen, Martin~J Wainwright, and Bin Yu.
\newblock Log-concave sampling: {M}etropolis-{H}astings algorithms are fast!
\newblock In \emph{Proceedings of the 31st Conference On Learning Theory},
  volume~75 of \emph{Proceedings of Machine Learning Research}, pages 793--797.
  PMLR, 2018.

\bibitem[Fu and Zhang(2017)]{fu2017cpsg}
Tianfan Fu and Zhihua Zhang.
\newblock {CPSG-MCMC}: {C}lustering-{B}ased {P}reprocessing method for
  {S}tochastic {G}radient {MCMC}.
\newblock In \emph{Proceedings of the 20th International Conference on
  Artificial Intelligence and Statistics}, volume~54 of \emph{Proceedings of
  Machine Learning Research}, pages 841--850. PMLR, 2017.

\bibitem[Gorham and Mackey(2017)]{gorham2017measuring}
Jackson Gorham and Lester Mackey.
\newblock Measuring {S}ample {Q}uality with {K}ernels.
\newblock In \emph{Proceedings of the 34th International Conference on Machine
  Learning}, volume~70 of \emph{Proceedings of Machine Learning Research},
  pages 1292--1301. PMLR, 2017.

\bibitem[Hastings(1970)]{hastings1970monte}
W.K. Hastings.
\newblock Monte {C}arlo sampling methods using {M}arkov chains and their
  applications.
\newblock \emph{Biometrika}, 57\penalty0 (1):\penalty0 97--109, 1970.

\bibitem[Kern and Gyorgy(2016)]{kern2016svrg}
Tam{\'a}s Kern and A~Gyorgy.
\newblock {SVRG}++ with {N}on-uniform {S}ampling.
\newblock In \emph{Proceedings of the 9th NIPS Workshop on Optimization for
  Machine Learning}, 2016.

\bibitem[Kingma and Ba(2016)]{adam2015}
Diederik~P. Kingma and Jimmy Ba.
\newblock Adam: {A} {M}ethod for {S}tochastic {O}ptimization.
\newblock In \emph{Proceedings of the 3rd International Conference on Learning
  Representations (ICLR)}, 2016.

\bibitem[Li et~al.(2021)Li, Wang, Zha, and Tao]{li2021improving}
Ruilin Li, Xin Wang, Hongyuan Zha, and Molei Tao.
\newblock Improving {S}ampling {A}ccuracy of {S}tochastic {G}radient {MCMC}
  {M}ethods via {N}on-uniform {S}ubsampling of {G}radients.
\newblock \emph{Discrete and Continuous Dynamical Systems - S}, 0, 2021.

\bibitem[Liu et~al.(2016)Liu, Lee, and Jordan]{liu2016kernelized}
Qiang Liu, Jason Lee, and Michael Jordan.
\newblock A {K}ernelized {S}tein {D}iscrepancy for {G}oodness-of-fit {T}ests.
\newblock In \emph{Proceedings of the 33rd International Conference on Machine
  Learning}, volume~48 of \emph{Proceedings of Machine Learning Research},
  pages 276--284. PMLR, 2016.

\bibitem[Liu et~al.(2020)Liu, Wu, and Mozafari]{liu2020adam}
Rui Liu, Tianyi Wu, and Barzan Mozafari.
\newblock Adam with {B}andit {S}ampling for {D}eep {L}earning.
\newblock In \emph{Advances in Neural Information Processing Systems},
  volume~33, pages 5393--5404, 2020.

\bibitem[Metropolis et~al.(1953)Metropolis, Rosenbluth, Rosenbluth, Teller, and
  Teller]{metropolis1953equation}
Nicholas Metropolis, Arianna~W Rosenbluth, Marshall~N Rosenbluth, Augusta~H
  Teller, and Edward Teller.
\newblock Equation of {S}tate {C}alculations by {F}ast {C}omputing {M}achines.
\newblock \emph{Journal of Chemical Physics}, 21\penalty0 (6):\penalty0
  1087--1092, 1953.

\bibitem[Neal(2011)]{neal2011mcmc}
Radford~M Neal.
\newblock M{CMC} using {H}amiltonian {D}ynamics.
\newblock In \emph{Handbook of Markov Chain Monte Carlo}, chapter~5. Chapman
  and Hall/CRC, 2011.

\bibitem[Nemeth and Fearnhead(2021)]{nemeth2021stochastic}
Christopher Nemeth and Paul Fearnhead.
\newblock Stochastic gradient {M}arkov chain {M}onte {C}arlo.
\newblock \emph{Journal of the American Statistical Association}, 116\penalty0
  (533):\penalty0 433--450, 2021.

\bibitem[Parisi(1981)]{parisi1981}
Giorgio Parisi.
\newblock Correlation functions and computer simulations.
\newblock \emph{Nuclear Physics B}, 180\penalty0 (3):\penalty0 378--384, 1981.

\bibitem[Robert and Casella(2004)]{robert2004}
Christian Robert and George Casella.
\newblock \emph{Monte Carlo Statistical Methods}.
\newblock Springer Science \& Business Media, 2nd edition, 2004.

\bibitem[Roberts and Rosenthal(1998)]{roberts1998optimal}
Gareth~O Roberts and Jeffrey~S Rosenthal.
\newblock Optimal scaling of discrete approximations to {L}angevin diffusions.
\newblock \emph{Journal of the Royal Statistical Society: Series B (Statistical
  Methodology)}, 60\penalty0 (1):\penalty0 255--268, 1998.

\bibitem[Roberts and Tweedie(1996)]{roberts1996exponential}
Gareth~O Roberts and Richard~L Tweedie.
\newblock Exponential convergence of {L}angevin distributions and their
  discrete approximations.
\newblock \emph{Bernoulli}, pages 341--363, 1996.

\bibitem[Salehi et~al.(2017)Salehi, Celis, and Thiran]{salehi2017stochastic}
Farnood Salehi, L~Elisa Celis, and Patrick Thiran.
\newblock Stochastic {O}ptimization with {B}andit {S}ampling.
\newblock \emph{arXiv preprint arXiv:1708.02544}, 2017.

\bibitem[Schmidt et~al.(2015)Schmidt, Babanezhad, Ahmed, Defazio, Clifton, and
  Sarkar]{schmidt15}
Mark Schmidt, Reza Babanezhad, Mohamed Ahmed, Aaron Defazio, Ann Clifton, and
  Anoop Sarkar.
\newblock {Non-{U}niform {S}tochastic {A}verage {G}radient {M}ethod for
  {T}raining {C}onditional {R}andom {F}ields}.
\newblock In \emph{Proceedings of the Eighteenth International Conference on
  Artificial Intelligence and Statistics}, volume~38 of \emph{Proceedings of
  Machine Learning Research}, pages 819--828. PMLR, 2015.

\bibitem[Schmidt et~al.(2017)Schmidt, Le~Roux, and Bach]{schmidt2017minimizing}
Mark Schmidt, Nicolas Le~Roux, and Francis Bach.
\newblock Minimizing {F}inite {S}ums with the {S}tochastic {A}verage
  {G}radient.
\newblock \emph{Mathematical Programming}, 162:\penalty0 83--112, 2017.

\bibitem[Teh et~al.(2016)Teh, Thiery, and Vollmer]{teh2016consistency}
Yee~Whye Teh, Alexandre~H Thiery, and Sebastian~J Vollmer.
\newblock Consistency and {F}luctuations {F}or {S}tochastic {G}radient
  {L}angevin {D}ynamics.
\newblock \emph{Journal of Machine Learning Research}, 17\penalty0
  (7):\penalty0 1--33, 2016.

\bibitem[Vollmer et~al.(2016)Vollmer, Zygalakis, and Teh]{vollmer2016non}
Sebastian~J. Vollmer, Konstantinos~C. Zygalakis, and Yee~Whye Teh.
\newblock Exploration of the ({N}on-){A}symptotic {B}ias and {V}ariance of
  {S}tochastic {G}radient {L}angevin {D}ynamics.
\newblock \emph{Journal of Machine Learning Research}, 17\penalty0
  (159):\penalty0 1--48, 2016.

\bibitem[Welling and Teh(2011)]{welling2011bayesian}
Max Welling and Yee~Whye Teh.
\newblock Bayesian {L}earning via {S}tochastic {G}radient {L}angevin
  {D}ynamics.
\newblock In \emph{Proceedings of the 28th International Conference on Machine
  Learning}, pages 681--688, 2011.

\bibitem[Zhang et~al.(2017)Zhang, Kjellstrom, and
  Mandt]{zhang2017determinantal}
Cheng Zhang, Hedvig Kjellstrom, and Stephan Mandt.
\newblock Determinantal {P}oint {P}rocesses for {M}ini-{B}atch
  {D}iversification.
\newblock In \emph{Proceedings of the 33rd Conference on Uncertainty in
  Artificial Intelligence (UAI)}, pages 1--13, 2017.

\bibitem[Zhao and Zhang(2014{\natexlab{a}})]{zhao2014accelerating}
Peilin Zhao and Tong Zhang.
\newblock Accelerating {M}inibatch {S}tochastic {G}radient {D}escent using
  {S}tratified {S}ampling.
\newblock \emph{arXiv preprint arXiv:1405.3080}, 2014{\natexlab{a}}.

\bibitem[Zhao and Zhang(2014{\natexlab{b}})]{zhao2014stochastic}
Peilin Zhao and Tong Zhang.
\newblock Stochastic {O}ptimization with {I}mportance {S}ampling.
\newblock \emph{arXiv preprint arXiv:1401.2753}, 2014{\natexlab{b}}.

\bibitem[Zhao and Zhang(2015)]{zhao2015stochastic}
Peilin Zhao and Tong Zhang.
\newblock Stochastic {O}ptimization with {I}mportance {S}ampling for
  {R}egularized {L}oss {M}inimization.
\newblock In \emph{Proceedings of the 32nd International Conference on Machine
  Learning}, volume~37 of \emph{Proceedings of Machine Learning Research},
  pages 1--9. PMLR, 2015.

\end{thebibliography}
\clearpage
\appendix

\section{Results from Section~\ref{sec:method}}
\subsection{Full derivation of the pseudo-variance} 
\label{sec:pseudovar}
The pseudo-variance $\tilde{g}^{(t)}$ is given by: 
\begin{align}
  \V\big(\tilde{g}^{(t)}\big) &= \E \bigg( \big\| \tilde{g}^{(t)} - g^{(t)} \big\|^2\bigg)  \\
                   &= \E \bigg (\big(\tilde{g}^{(t)} - g^{(t)}\big)^T \big(\tilde{g}^{(t)} - g^{(t)}\big) \bigg) \\
                   &= \sum_{j=1}^d \E\bigg(\big(\tilde{g}^{(t)}_j - g_j^{(t)}\big)^2\bigg) \text{\, \,\, \, \, (decompose expectation over $d$ parameters)}\\
                   &= \sum_{j=1}^d  \E\bigg(\big(\tilde{g}^{(t)}_j - \E\big[\,\tilde{g}^{(t)}_j\big]\big)^2\bigg) \\
                   &= \sum_{j=1}^d \Var\big(\tilde{g}^{(t)}_j\big) \\
                   &= \tr\bigg(\Cov \big(\tilde{g}^{(t)}\big)\bigg).
\end{align}  

\subsection{Proof of Lemma~\ref{lemma3.1}}
\begin{proof}
From Section \ref{sec:pseudovar}, we know that 
\begin{align} \label{eq:componentsum1}
 \mathbb{V}\big(\tilde{g}^{(t)} \big) =  \E \Bigg[ \big\| \tilde{g}^{(t)} - g^{(t)} \big\|^2\Bigg] = \sum_{j=1}^d \Var\big(\tilde{g}_j^{(t)}\big).\end{align}
 Taking expectations with respect to $\mathbf{p}^{(t)}$, we know that the $j$-th component of the sum in Eq.~\eqref{eq:componentsum1} is given by: 
 \begin{align*}
   \Var\big(\tilde{g}_j^{(t)} \big) &= \Var \bigg( \nabla_j f_0\big(\theta^{(t)}\big) + \frac{1}{n} \sum_{i \in \mathcal{S}^t} \frac{1}{p_{i}^t} \nabla_j f_i \big(\theta^{(t)}\big) \bigg) = \Var \bigg(\frac{1}{n} \sum_{i \in \mathcal{S}^t} \frac{1}{p_{i}^t} \nabla_j f_i \big(\theta^{(t)}\big)\bigg). 
   \end{align*}
 Given that all stochastic gradient estimators are unbiased with the same mean, we know that minimising $\mathbb{V}\big(\tilde{g}^{(t)} \big)$ with respect to $\mathbf{p}^{(t)}$ is equivalent to minimising the component-wise sum of second moments, $\sum_{j=1}^d  \E \Bigg( \bigg(\frac{1}{n} \sum_{i \in \mathcal{S}^t} \frac{1}{p_{i}^t} \nabla_j f_i \big(\theta^{(t)}\big) \bigg)^2 \Bigg).$
 
 So, for $j = 1, \ldots, d$, we consider 
 \begin{align*}
   \E \Bigg( \bigg(\frac{1}{n} \sum_{i \in \mathcal{S}^t} \frac{1}{p_{i}^t} \nabla_j f_i \big(\theta^{(t)}\big) \bigg)^2 \Bigg) 
  &= \frac{1}{n^2} \, \E \Bigg(  \sum_{i \in \mathcal{S}^t} \bigg(\frac{1}{p_{i}^t} \nabla_j f_i \big(\theta^{(t)}\big) \bigg)^2   +  \sum_{i \in \mathcal{S}^t}  \sum_{k \in \mathcal{S}^t, \, i \neq k} \frac{1}{p_{i}^t} \cdot \frac{1}{p_{k}^t} \cdot \nabla_j f_i \big(\theta^{(t)}\big)  \cdot   \nabla_j f_k \big(\theta^{(t)}\big) \Bigg)  \\ 
   &= \frac{1}{n} \, \E \Bigg(  \bigg(\frac{1}{p_{i}^t} \nabla_j f_i \big(\theta^{(t)}\big)\bigg)^2  \Bigg)  + \frac{n(n-1)}{n^2}  \E \Bigg( \frac{1}{p_{i}^t}  \nabla_j f_i \big(\theta^{(t)}\big)  \Bigg)^2  
 \\   &=\frac{1}{n}\sum_{i=1}^N \frac{1}{p_{i}^t} \big[\nabla_j f_i \big(\theta^{(t)}\big) \big]^2  + \frac{n-1}{n} \Bigg( \sum_{i=1}^N  \nabla_j f_i \big(\theta^{(t)}\big)  \Bigg)^2   \\ 
&= \frac{1}{n}\sum_{i=1}^N \frac{1}{p_{i}^t} \big[\nabla_j f_i \big(\theta^{(t)}\big) \big]^2  + C_j, 
\end{align*} 
where $C_j$ is a constant that does not depend on $\mathbf{p}^{(t)}$. Adding up over all components, we see that,
\begin{align*}
\sum_{j=1}^d  \E \Bigg( \bigg(\frac{1}{n} \sum_{i \in \mathcal{S}^t} \frac{1}{p_{i}^t} \nabla_j f_i \big(\theta^{(t)}\big) \bigg)^2 \Bigg) &=  \sum_{j=1}^d \bigg[\frac{1}{n} \sum_{i=1}^N  \frac{1}{p_i^t}   \big[\nabla_j f_i \big(\theta^{(t)}\big)  \big]^2 + C_j \bigg] \\ &= \frac{1}{n} \sum_{i=1}^N \frac{1}{p_i^t} \| \nabla f_i \big(\theta^{(t)}\big) \|^2 + C'.
\end{align*}
Therefore, 
\begin{align*}
\min_{\mathbf{p}^{(t)}, \, p_{i}^t \in [0,1], \sum_i p_{i}^t = 1} \mathbb{V}\big(\tilde{g}^{(t)} \big) \iff \min_{\mathbf{p}^{(t)}, \, p_{i}^t \in [0,1], \sum_i p_{i}^t = 1}  \frac{1}{n}\sum_{i=1}^N \frac{1}{p_{i}^t} \big\|\nabla f_i \big(\theta^{(t)}\big) \big\|^2.
\end{align*} 

The minimisation problem of interest is
\begin{align*}
 \min_{\mathbf{p}^{(t)}, \, p_{i}^t \in [0,1], \sum_i p_{i}^t = 1}  \frac{1}{n}\sum_{i=1}^N \frac{1}{p_{i}^t} \big\|\nabla f_i \big(\theta^{(t)}\big) \big\|^2.
\end{align*}

We know that via the Cauchy-Schwarz inequality\footnote{The Cauchy-Schwarz inequality states that for $d$-dimensional real vectors $\mathbf{u}, \mathbf{v} \in \mathbb{R}^d$, of all inner product space it is true that $$|\langle \mathbf{u}, \mathbf{v}\rangle| \leq \langle \mathbf{u},\mathbf{u}\rangle \cdot \langle \mathbf{v},  \mathbf{v}\rangle$$ Furthermore, the equality holds only when either $\mathbf{u}$ or $\mathbf{v}$ is a multiple of the other.}, 
\begin{align*}
   \frac{1}{n}\sum_{i=1}^N \frac{1}{p_{i}^t} \big\|\nabla f_i \big(\theta^{(t)}\big)  \big\|^2
  &= \frac{1}{n} \Bigg(\sum_{i=1}^N \frac{1}{p_{i}^t}  \big\|\nabla f_i \big(\theta^{(t)}\big)  \big\|^2 \Bigg) \bigg(\sum_{i=1}^N p_{i}^t \bigg) \\
  &\geq \frac{1}{n} \Bigg( \sum_{i=1}^N \sqrt{\frac{1}{p_{i}^t} \big\|\nabla f_i \big(\theta^{(t)}\big) \big\|^2 p_{i}^t } \,\Bigg)^2 \\
  &= \frac{1}{n} \bigg (\sum_{i=1}^N \big\|\nabla f_i \big(\theta^{(t)}\big)  \big\| \bigg)^2.
\end{align*}
The equality is only obtained when there exists a constant $c \in \mathbb{R}$ such that
\begin{align*}
  \begin{pmatrix}
    (p_{1}^t)^{-1} \big\|\nabla f_1 \big(\theta^{(t)}\big) \big\|^2 \\
   (p_{2}^t)^{-1} \big\|\nabla f_2 \big(\theta^{(t)}\big)  \big\|^2 \\
    \vdots \\
   (p_{N}^t)^{-1} \big\|\nabla f_N \big(\theta^{(t)}\big) \big\|^2 
  \end{pmatrix} = c
  \begin{pmatrix}
    p_{1}^t \\
    p_{2}^t \\
    \vdots \\
    p_{N}^t
  \end{pmatrix},
\end{align*} which is equivalent to writing
\begin{align*}
  (p_{1}^t)^{-2}\big\|\nabla f_1 \big(\theta^{(t)}\big) \big\|^2 = (p_{2}^t)^{-2}\big\|\nabla f_2 \big(\theta^{(t)}\big)  \big\|^2 = \ldots = (p_{N}^t)^{-2} \big\|\nabla f_N \big(\theta^{(t)}\big) \big\|^2.
\end{align*} Under this constraint, Problem~\eqref{eq:min3} is minimised. We can therefore conclude that the optimal weights which minimise the pseudo-variance are,
\begin{align*}  p_{i}^{t} = \frac{\| \nabla f_i \big(\theta^{(t)}\big)\|}{\sum_{k=1}^N \| \nabla f_k \big(\theta^{(t)}\big)  \| } \enspace \text{ for } i=1, \ldots, N. \end{align*}
\end{proof}

\subsection{Proof of Lemma~\ref{lemma3.2}}
\begin{proof}
By the similar argument to that used for Lemma~\ref{lemma3.1}, we can show that 
\begin{align*}
\min_{\mathbf{p}^{(t)}, \, p_{i}^t \in [0,1], \sum_i p_{i}^t = 1} \mathbb{V}\big(\tilde{g}^{(t)} \big) \iff \min_{\mathbf{p}^{(t)}, \, p_{i}^t \in [0,1], \sum_i p_{i}^t = 1}  \frac{1}{n}\sum_{i=1}^N \frac{1}{p_{i}^t} \big\|\nabla f_i \big(\theta^{(t)}\big) - \nabla f_i(\hat{\theta}) \big\|^2.
\end{align*} 

Once again, using the Cauchy-Schwarz inequality allows us to see that the optimal weights which minimise the pseudo-variance are, 
\begin{align*}  p_{i}^{t} = \frac{\| \nabla f_i \big(\theta^{(t)}\big) - \nabla f_i(\hat{\theta})) \|}{\sum_{k=1}^N \| \nabla f_k \big(\theta^{(t)}\big) - \nabla f_k(\hat{\theta}) \| } \enspace \text{ for } i=1, \ldots, N. 
\end{align*}
\end{proof}

\subsection{Deriving approximate weights for the control variates gradient} \label{sec:weights_analytical}
The minimisation problem of interest is 
\begin{align*}
 \min_{\mathbf{p}^{(t)}, \, p_{i}^t \in [0,1], \sum_i p_{i}^t = 1} \mathbb{E}_\theta \bigg[ \frac{1}{n}\sum_{i=1}^N \frac{1}{p_{i}^t} \big\|\nabla f_i \big(\theta^{(t)}\big) - \nabla f_i(\hat{\theta}) \big\|^2 \bigg].
 \end{align*} So, 
\begin{align*} 
\mathbb{E}_\theta \bigg[ \frac{1}{n}\sum_{i=1}^N \frac{1}{p_{i}^t} \big\|\nabla f_i \big(\theta^{(t)}\big) - \nabla f_i(\hat{\theta}) \big\|^2 \bigg] 
&= \frac{1}{n} \sum_{i=1}^N \frac{1}{p_{i}^t} \mathbb{E}_\theta \bigg[ \big\|\nabla f_i \big(\theta^{(t)}\big) - \nabla f_i(\hat{\theta}) \big\|^2 \bigg] \text{\,\, (linearity of $\mathbb{E}_{\theta}$) }\\
&= \bigg(\frac{1}{n}\bigg)\bigg(\sum_{i=1}^N \frac{1}{p_{i}^t} \mathbb{E}_\theta \bigg[ \big\|\nabla f_i \big(\theta^{(t)}\big) - \nabla f_i(\hat{\theta}) \big\|^2 \bigg] \bigg)\bigg(\sum_{i=1}^N p_i^t\bigg) \\
&\geq \frac{1}{n}\bigg(\sum_{i=1}^N\sqrt{ \frac{1}{p_i^t }\, \mathbb{E}_\theta \bigg[ \big\|\nabla f_i \big(\theta^{(t)}\big) - \nabla f_i(\hat{\theta}) \big\|^2 \bigg] p_i^t \, }\bigg) \text{\,\, (Cauchy-Schwarz) } \\
&= \frac{1}{n} \bigg(\sum_{i=1}^N \sqrt{\mathbb{E}_\theta \bigg[ \big\|\nabla f_i \big(\theta^{(t)}\big) - \nabla f_i(\hat{\theta}) \big\|^2 \bigg]} \, \, \bigg)^2.
\end{align*}
The problem is minimised when 
\begin{align*}
  p_i^t \propto \sqrt{\mathbb{E}_\theta \bigg[ \big\|\nabla f_i \big(\theta^{(t)}\big) - \nabla f_i(\hat{\theta}) \big\|^2 \bigg]} \text{ for } i=1,\ldots,N. 
\end{align*}
Let's assume that $\theta \mathrel{\dot\sim} \mathcal{N}(\hat{\theta}, \hat{\Sigma})$ at stationarity, where $\hat{\Sigma} = -H(\hat{\theta})^{-1}$. Using a first-order Taylor expansion about $\hat{\theta}$, 
\begin{align*}
  \big\|\nabla f_i \big(\theta^{(t)}\big) - \nabla f_i(\hat{\theta}) \big\|^2 \approx \| \nabla f_i(\hat{\theta}) + \nabla^2f_i(\hat{\theta})(\theta^{(t)}-\hat{\theta}) - \nabla f_i(\hat{\theta})\big\|^2 = \big\|  \nabla^2f_i(\hat{\theta})(\theta^{(t)}-\hat{\theta})\big\|^2.
\end{align*}
We know that $(\theta - \hat{\theta}) \mathrel{\dot\sim} \mathcal{N}(\mathbf{0}, \hat{\Sigma})$. So,
$$ \nabla^2f_i(\hat{\theta})(\theta^{(t)}-\hat{\theta}) \mathrel{\dot\sim} \mathcal{N}\big(\mathbf{0}, \nabla^2f_i(\hat{\theta})\hat{\Sigma}\nabla^2f_i(\hat{\theta})^T\big).$$
Then,
\begin{align*}
  \mathbb{E}_\theta \bigg[ \big\|\nabla f_i \big(\theta^{(t)}\big) - \nabla f_i(\hat{\theta}) \big\|^2 \bigg] &\approx \mathbb{E}_\theta \bigg[ \big\|\nabla^2f_i(\hat{\theta})(\theta^{(t)}-\hat{\theta}) \big\|^2 \bigg] \\&= \text{tr}\bigg( \text{Cov}\bigg(\nabla^2f_i(\hat{\theta})\big(\theta^{(t)}-\hat{\theta}\big) \bigg) \bigg) \\&\approx \text{tr}\bigg(  \nabla^2f_i(\hat{\theta})\,\hat{\Sigma}\,\nabla^2f_i(\hat{\theta})^T \bigg).
\end{align*}
So, we can set
\begin{align*}
  p_i \propto \sqrt{\text{tr}\bigg( \nabla^2f_i(\hat{\theta})\,\hat{\Sigma}\,\nabla^2f_i(\hat{\theta})^T\bigg)} \text{ for } i=1,\ldots,N.
\end{align*}
\subsection{Proof of Lemma \ref{lemma3.3}}
\begin{proof}
We know that the pseudo-variance can be decomposed into 
\begin{align*}
\V(\tilde{g}) &= \E[\|\tilde{g} - g\|^2] \\
&= \frac{1}{n} \sum_{i=1}^N \frac{1}{p_i^t} \| \nabla f_i \big(\theta^{(t)}\big) - \nabla f_i(\hat{\theta})\|^2 -\underbrace{\frac{1}{n}\bigg\|\sum_{i=1}^N \big[\grad_j f_i\big(\theta^{(t)}\big) - \nabla_j f_i(\hat{\theta}) \big] \bigg\|^2}_{\geq 0} \\
&\leq \frac{1}{n} \sum_{i=1}^N \frac{1}{p_i^t} \| \nabla f_i \big(\theta^{(t)}\big) - \nabla f_i(\hat{\theta})\|^2.
\end{align*} 
Under Assumption~\ref{assumption1}, we know that
\begin{align*}
    \V(\tilde{g}) \leq \frac{1}{n} \|\theta^{(t)} - \hat{\theta} \|^2 \bigg(\sum_{i=1}^N \frac{L_i^2}{p_i^t}\bigg)  \,.
\end{align*}
\end{proof}

\section{Pseudocode for algorithms}
\label{sec:alg-section}

\label{sec:app_algo}
\begin{algorithm}[H]
  \caption{SGLD-CV} \label{alg:sgldcv}
  \begin{algorithmic}[1]
    \STATE{Input: initialise $\theta^{(1)} = \hat{\theta}$, gradients $\nabla f_i \big(\hat{\theta})$, batch size $n$, step-size $\epsilon$.}
   \FOR{$t = 1, 2, \ldots, T$}
   \STATE{Sample indices $S^t \subset \{1,\ldots,N\}$ with or without replacement.}
   \STATE{Calculate $\hat{g}_{cv}^{(t)}$ in Eq.~\eqref{eq:ghat_cv}.}
   \STATE{Update parameters according to Eq.~\eqref{eq:sgldupdate}. }
    \ENDFOR 
    \RETURN $\theta^{(T+1)}$ 
 \end{algorithmic}
\end{algorithm}

\begin{algorithm}[ht]
  \caption{SGLD with preferential subsampling (SGLD-PS)} \label{alg:sgldps}
  \begin{algorithmic}[1]
    \STATE{Input: initialise $\theta^{(1)}$, weights $ \mathbf{p}^{(1)}$, batch size $n$, step-size  $\epsilon$.}
   \FOR{$t = 1, 2, \ldots, T$}
   \STATE{Update $\mathbf{p}^{(t)}$.}
   \STATE{Sample indices $S^t$ according to $\mathbf{p}^{(t)}$ with replacement.}
   \STATE{Calculate $\tilde{g}^{(t)}$ using Eq.~\eqref{eq:gtilde}.}
   \STATE{Update parameters  $\theta^{(t+1)} \leftarrow \theta^{(t)} - \frac{\epsilon}{2} \cdot  \tilde{g}^{(t)} + \N_d(0, \epsilon_t I_{d \times d})$ }
    \ENDFOR 
    \RETURN $\theta^{(T+1)}$ 
 \end{algorithmic}
\end{algorithm}

\begin{algorithm}[H]
 \caption{SGLD-CV with preferential subsampling (SGLD-CV-PS)}
 \label{alg:sgldcvps}
 \begin{algorithmic}[1]
  \STATE{Input: initialise $\theta^{(1)}$ close to the mode $\hat{\theta}$, gradients $\nabla f_i \big(\hat{\theta})$, weights $ \mathbf{p}^{(1)}$, batch size $n$, step-size  $\epsilon$.}
  \FOR{$t = 1, 2, \ldots, T$}
  \STATE{Update $\mathbf{p}^{(t)}$.}
  \STATE{Sample indices $S^t$ according to $\mathbf{p}^{(t)}$ with replacement.}
  \STATE{Calculate $\tilde{g}^{(t)}$ using Eq.~\eqref{eq:gtilde_cv}.}
   \STATE{Update parameters  $\theta^{(t+1)} \leftarrow \theta^{(t)} - \frac{\epsilon}{2} \cdot  \tilde{g}^{(t)} + \N_d(0, \epsilon_t I_{d \times d})$ }
   \ENDFOR 
\RETURN $\theta^{(T+1)}$ 
\end{algorithmic}
\end{algorithm}

%%%%%%%%%%%%%%%%%%%%%%%%%%%%%%%%%%%%%%%%%%%%%%%%%

\section{Model details}
\label{sec:app_models}
\subsection{Bivariate Gaussian}
\subsubsection{Model specification}
We want to simulate independent data from:
$$X_i|\theta \sim \mathcal{N}_2(\theta, \Sigma_x) \, \, \text{ for } i=1,\ldots,N.$$ 

It is assumed that that $\theta$ is unknown and $\Sigma_x$ is known. 
The likelihood for a single observation is given by:
$$p(x_i | \theta) = \frac{1}{\sqrt{(2\pi)^2 |\Sigma_x|}} \exp \Bigg( -\frac{1}{2} (x_i-\theta)^T\Sigma_x^{-1}(x_i-\theta)\Bigg).$$

The likelihood function for $N$ observations is
\begin{align*} p(\mathbf{x} | \theta) =  \prod_{i=1}^N \frac{1}{\sqrt{(2\pi)^2 |\Sigma_x|}} \exp \Bigg( -\frac{1}{2} (x_i-\theta)^T\Sigma_x^{-1}(x_i-\theta)\Bigg) \\ \propto |\Sigma_x|^{-\frac{N}{2}}\exp\Bigg(-\frac{1}{2} \sum_{i=1}^N (x_i - \theta)^T\Sigma_x^{-1} (x_i - \theta) \Bigg).\end{align*}

The log-likelihood is a quadratic form in $\theta$, and therefore the conjugate prior distribution for $\theta$ is the multivariate normal distribution. The conjugate prior for $\theta$ is set to be 
$$\theta \sim \mathcal{N}_2(\mu_0, \Lambda_0). $$

The conjugate posterior that we are ultimately trying to simulate from using SGLD is known to be:
$$\pi (\theta | \mathbf{x}) \propto \exp \Bigg(-\frac{1}{2} (\theta - \mu_1)^T \Lambda_1^{-1} (\theta - \mu_1) \Bigg) \overset{D}{=} \mathcal{N}_2(\mu_1, \Lambda_1),$$
where
$$\mu_1 = (\Lambda_0^{-1} + N\Sigma_x^{-1})^{-1} (\Lambda_0^{-1} \mu_0 + N\Sigma_x^{-1} \bar{x}), $$
and 
$$\Lambda_1^{-1} = \Lambda_0^{-1} + N \Sigma_x^{-1}. $$ 
\subsubsection{Model gradient}
For the prior, 
\begin{align*}
    \log p(\theta) = -\frac{1}{2}(\theta - \mu_0)^T \Lambda_0^{-1} (\theta - \mu_0) \end{align*}
Therefore, 
\begin{align*}
   \nabla f_0(\theta) =  - \nabla \log p(\theta) = \Lambda_0^{-1} (\theta - \mu_0).
\end{align*}
We know that for $i \in \{1, \ldots, N\}$,
$$f_i(\theta) = - \log p(x_i | \theta)  =  \frac{1}{2} (x_i - \theta)^T \Sigma_x^{-1} (x_i - \theta) + \text{constant}$$
Therefore,
$$\grad f_i(\theta) = \Sigma_x^{-1}(\theta - x_i) \text{\, and \,} \nabla^2f_i(\theta) = \Sigma_x^{-1}.$$
%Using results from \cite{dwivedi2018log}, we know that $f_i(\theta)$ is $L_i$-smooth with $L_i = \lambda^{-1}_\text{max}(\Sigma_x)$. 

\subsubsection{Synthetic data}
To generate the synthetic data, $N$ data points are drawn from the model with $\theta = \begin{pmatrix} 0 \\ 1 \end{pmatrix}$ and $\Sigma_x = \begin{pmatrix} 1\times10^5 & 6\times10^4 \\ 6\times10^4 & 2\times10^5 \end{pmatrix}$. The hyperparameters of the prior are $\mu_0 = \begin{pmatrix} 0 \\ 0 \end{pmatrix}$ and $\Lambda_0 = \begin{pmatrix} 1\times10^3 & 0 \\ 0 & 1\times10^3 \end{pmatrix}$.
Synthetic datasets of sizes $N=10^3$ and $N=10^4$ were generated for use in Figures~\ref{fig:grad_comp}(a) and \ref{fig:sgld_comp}(a) respectively. 
\subsection{Logistic regression}
\subsubsection{Model specification}
Suppose we have data $x_1, \dots, x_N$ of dimension $d$ taking values in $\mathbb R^{d}$, where each $x_i = (1, x_{i1}, \ldots, x_{ip})^T$ ($d=p+1$). Let us suppose that we also have the corresponding response variables $y_1, \dots, y_N$ taking values in $\{0, 1\}$. Then, a logistic regression model with parameters $\theta=(\beta_0,\beta_1, \ldots, \beta_{p})$ representing the coefficients $\beta_j$ for $j=1,\ldots,p$ and bias $\beta_0$ will have likelihood
$$ p( X, y | \theta ) = \prod_{i=1}^N \left[ \frac{1}{1 + e^{-\theta^T x_i}} \right]^{y_i} \left[ 1 - \frac{1}{1 + e^{-\theta^T x_i}} \right]^{1-y_i}$$

The prior for $\theta$ is set to be $\theta \sim \mathcal{N}_d(\mu_0, \Lambda_0). $  The hyperparameters of the prior are $\mu_0 = (0, \ldots,0)^T$ and $\Lambda_0 = \mbox{diag}(10,d)$.
\subsubsection{Model gradient and Hessian}
For the prior, 
\begin{align*}
    \log p(\theta) = -\frac{1}{2}(\theta - \mu_0)^T \Lambda_0^{-1} (\theta - \mu_0) = -\frac{1}{2}\theta^T \Lambda_0^{-1} \theta.
\end{align*}

Therefore, 
\begin{align*}
    \nabla f_0(\theta) =  - \log p(\theta) = \Lambda_0^{-1} \theta. 
\end{align*}
We know that $i \in \{1, \ldots, N\}$, the log-density 
\begin{align*} 
   \log p(y_i | x_i, \theta) &= y_i \log\bigg( \frac{1}{1+ \exp(-\theta^T x_i)}\bigg) + (1-y_i)\log \bigg(1- \frac{1}{1+ \exp(-\theta^T x_i)}\bigg) \\
    &=y_i \log\bigg( \frac{1}{1+ \exp(-\theta^T x_i)}\bigg) + (1-y_i)\log \bigg(\frac{\exp(-\theta^Tx_i)}{1+ \exp(-\theta^T x_i)}\bigg) \\
    &= y_i \log \bigg(\frac{1}{1+ \exp(-\theta^T x_i)}\cdot \frac{1+ \exp(-\theta^T x_i)}{\exp(-\theta^T x_i)} \bigg) + \log \bigg(\frac{\exp(-\theta^Tx_i)}{1+ \exp(-\theta^T x_i)}\bigg) \\
    &= y_i \log(\exp(\theta^Tx_i)) +  \log \bigg(\frac{1}{1+ \exp(\theta^T x_i)}\bigg) \\
    &= y_i \theta^Tx_i \, - \log(1+ \exp(\theta^T x_i))
\end{align*}
Therefore,
\begin{align*}
    f_i(\theta) = -  \log p(y_i | x_i, \theta) = \log(1+ \exp(\theta^T x_i)) - y_i \theta^Tx_i, 
\end{align*}
 and the corresponding gradient vector is 
\begin{align*}
    \nabla f_i(\theta) = \frac{\exp(\theta^T x_i)}{1+ \exp(\theta^T x_i)} \cdot x_i  - y_i \cdot x_i =  \frac{1}{1+ \exp(-\theta^T x_i)} \cdot x_i - \, y_i \cdot x_i 
\end{align*}
The corresponding Hessian is
\begin{align*}
    \nabla^2 f_i(\theta) =  \frac{\exp(-\theta^T x_i)}{(1+ \exp(-\theta^T x_i))^2} \cdot x_ix_i^T \,.
\end{align*}

We know that for the function $h(a) = \log(1+\exp(-a)), h''(a) = \frac{\exp(-a)}{(1+exp(-a))^2} \leq \frac{1}{4}$. Therefore, 

\begin{align*}
    \nabla^2 f_i(\theta) =  \frac{\exp(-\theta^T x_i)}{(1+ \exp(-\theta^T x_i))^2} \cdot x_ix_i^T \preceq \frac{1}{4} x_ix_i^T. 
\end{align*}

Using results from \cite{durmus2019high} and \cite{dwivedi2018log}, we know that $f_i(\theta)$ is $L_i$-continuous with $L_i = \frac{1}{4} \lambda_{\text{max}}(x_ix_i^T)$.

\subsubsection{Synthetic data}
We used the Python module \texttt{sklearn} to produce our synthetic classification data with four features ($d=5$). We have generated $N$ training data points and $N_{test} = 0.5N$ test data points. 

Two types of synthetic data were generated:
\begin{enumerate}
    \item Balanced classification data, where 50\% of instances have $y_i=1$. Synthetic datasets of sizes $N=10^3$ and $N=10^4$ were used for Figures~\ref{fig:grad_comp}(b) and \ref{fig:asgldcv_synth} respectively. 
    \item Highly imbalanced classification data, where 95\% of the instances have $y_i =1$. Synthetic data of size $N=10^3$ was used for Figure~\ref{fig:grad_comp}(c). 
\end{enumerate}

\subsubsection{Real data}
We used the covertype dataset \citep{blackard1998} for Figures~\ref{fig:sgldcv_comp}(b)-(c) and \ref{fig:sgld_comp}(b). The covertype dataset contains 581,012 instances and 54 features. In particular, we have used a transformed version of this dataset that is available via the LIBSVM repository\footnote{\url{https://www.csie.ntu.edu.tw/~cjlin/libsvmtools/datasets/binary.html}}. We split the covertype dataset into training and test sets with 75\% and 25\% of the instances respectively.
%%%%%%%%%%%%%%%%%%%%%%%%%%%%%%%%%%%%%%%%%%%%%%%%%

\subsection{Linear regression}
\subsubsection{Model specification}
Suppose we have data $x_1, \dots, x_N$ of dimension $d$ taking values in $\mathbb R^{d}$, where each $x_i = (1, x_{i1}, \ldots, x_{ip})^T$ ($d=p+1$). Let us suppose that we also have the corresponding response variables $y_1, \dots, y_N$ taking values on the real line. 

We define the following linear regression model,
$$ y_i = x_i^T \theta + \eta_i, \enspace \eta_i \sim \mathcal{N}(0, 1)$$
with parameters $\theta=(\beta_0,\beta_1, \ldots, \beta_{p})$ representing the coefficients $\beta_j$ for $j=1,\ldots,p$ and bias $\beta_0$. The regression model will thus have likelihood
$$ p( X, y | \theta ) = \prod_{i=1}^N \left[ \frac{1}{\sqrt{2\pi}} \exp\bigg(-\frac{1}{2}(y_i - x_i^T \theta)^2\bigg) \right].$$

The prior for $\theta$ is set to be $\theta \sim \mathcal{N}_d(\mu_0, \Lambda_0). $ The hyperparameters of the prior are $\mu_0 = (0, \ldots,0)^T$ and $\Lambda_0 = \mbox{diag}(10,d)$.
\subsubsection{Model gradient and Hessian}
\begin{align*}
    \log p(\theta) = -\frac{1}{2}(\theta - \mu_0)^T \Lambda_0^{-1} (\theta - \mu_0) = -\frac{1}{2}\theta^T \Lambda_0^{-1} \theta.
\end{align*}

Therefore, 
\begin{align*}
   \nabla f_0(\theta) =  - \nabla \log p(\theta) = \Lambda_0^{-1} \theta. 
\end{align*}

We know that $i \in \{1, \ldots, N\}$, the log-density 
\begin{align*} 
   \log p(y_i | x_i, \theta) = -\frac{1}{2} (y_i - x_i^T\theta)^2 - \frac{1}{2}\log(2\pi) 
\end{align*}
Therefore,
\begin{align*}
    f_i(\theta) = -  \log p(y_i | x_i, \theta) = \frac{1}{2} (y_i - x_i^T\theta)^2 + \frac{1}{2}\log(2\pi)  
\end{align*}
 and the corresponding gradient vector is 
\begin{align*}
    \nabla f_i(\theta) = - (y_i - x_i^T\theta)\cdot x_i.
\end{align*}
The corresponding Hessian is
\begin{align*}
    \nabla^2 f_i(\theta) = x_ix_i^T \,.
\end{align*}

Using results from \citet{dwivedi2018log}, we know that $f_i(\theta)$ is $L_i$-continuous with $L_i = \lambda_{\text{max}}(x_ix_i^T)$.
\subsubsection{Real data}
We used the CASP\footnote{\url{https://archive.ics.uci.edu/ml/datasets/Physicochemical+Properties+of+Protein+Tertiary+Structure}} dataset from the UCI Machine Learning repository for Figures~\ref{fig:sgldcv_comp}(a) and \ref{fig:asgldcv_casp}. The CASP dataset contains 45,730 instances and 9 features.

%%%%%%%%%%%%%%%%%%%%%%%%%%%%%%%%%%%%%%%%%%%%%%%%%
\section{Computational costs for the SGLD-CV-PS approximate subsampling weights}

\label{sec:app_comp_cost}
Recall that the optimal subsampling weights in Eq.~\eqref{eq:optp2} can be approximated by the following scheme,
\begin{align*} 
  p_i \propto \sqrt{\text{tr}\bigg( \nabla^2f_i(\hat{\theta})\,\hat{\Sigma}\,\nabla^2f_i(\hat{\theta})^T\bigg)} \text{ for } i=1,\ldots,N.
\end{align*}
Here, $\nabla^2f_i(\cdot)$ is the Hessian matrix of $f_i(\cdot)$ and $\hat{\Sigma}$ is the covariance matrix of the Gaussian approximation to the target posterior centred at the mode. 

In practice, these weights should be evaluated as a one-off preprocessing step before the SGLD-CV-PS chain is run. We now assess the total computational cost associated with this preprocessing step. 
\begin{itemize}
    \item The hessians of each $f_i(\cdot)$ need to evaluated at the mode. For each log-density function, this step costs $O(d^2)$.
    \item The covariance matrix of the Gaussian approximation of the posterior needs to calculated once. This involves inverting the observed information matrix at a cost of $O(d^3)$.
    \item The cost of multiplying three $d \times d$ square matrices is $O(d^3)$
    \item The cost of calculating the trace of a $d \times d$ matrix is $O(d)$.
    \item The cost of calculating the square root of a scalar is $O(1)$.
\end{itemize}
Therefore, the cost of calculating these weights for all $N$ data points is $O(Nd^3)$. As such, we recognise that there will be limits to where the SGLD-CV-PS algorithm can be used. The SGLD-CV-PS has been implemented with success on the covertype dataset \cite{blackard1998} (with 54 features and 581,012 instances and 54 features) in Section~\ref{sec:exp_2}. We recommend that this method is not implemented for models with more than $d>60$ parameters in practice. 

%%%%%%%%%%%%%%%%%%%%%%%%%%%%%%%%%%%%%%%%%%%%%%%%%
\section{Numerical experiment set-up}

\label{sec:app_exp_setup}
\subsection{Step-size selection}
\begin{figure*}[h]
\begin{center}
\begin{minipage}[c]{.33\textwidth}
\centering 
\begin{minipage}[c]{\textwidth}
  \centering
\includegraphics[width=\textwidth]{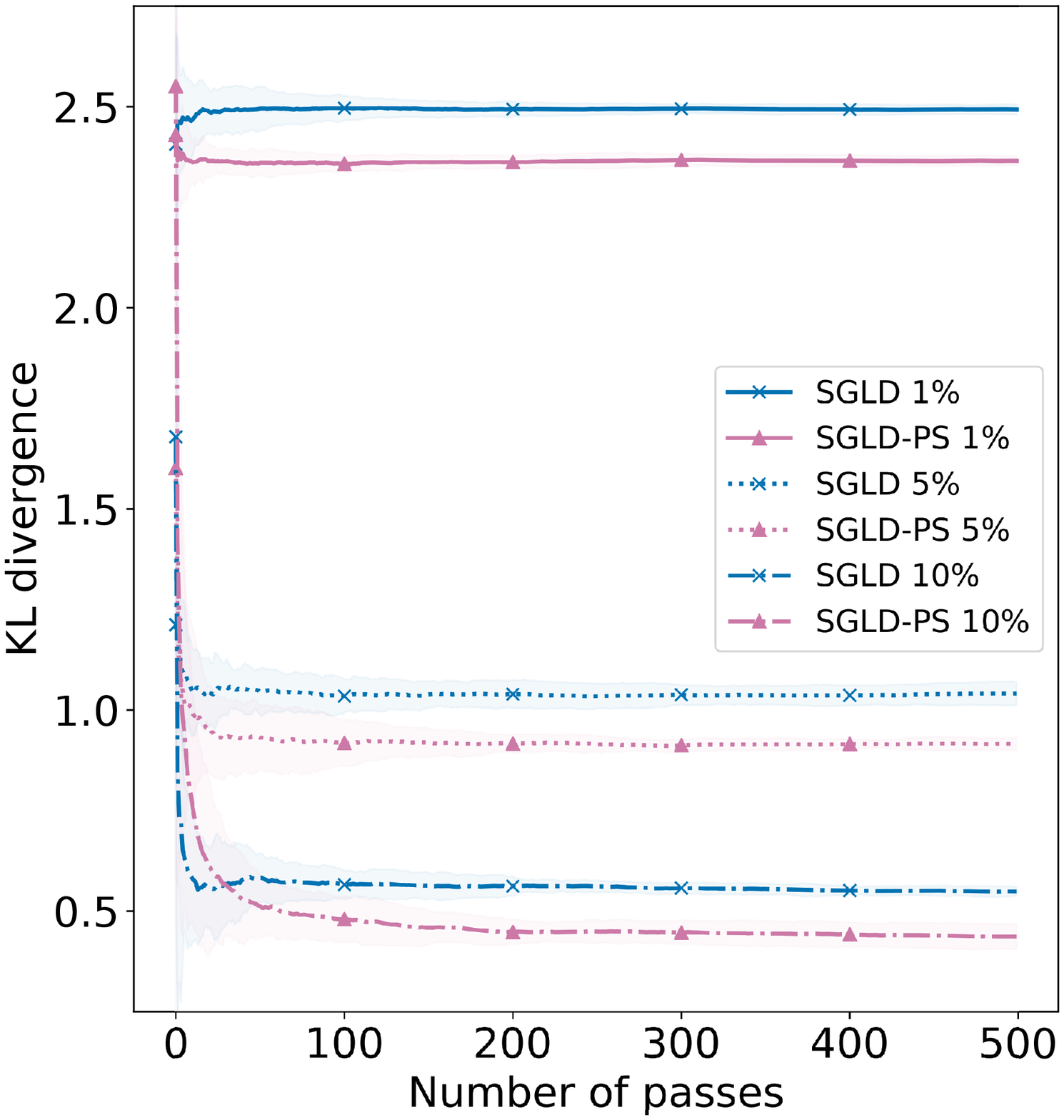} 
\vspace{-10pt}
\end{minipage}
(a)
\end{minipage}  \hspace{15pt}
\begin{minipage}[c]{.33\textwidth}
\centering 
\begin{minipage}[c]{\textwidth}
  \centering
\includegraphics[width=\textwidth]{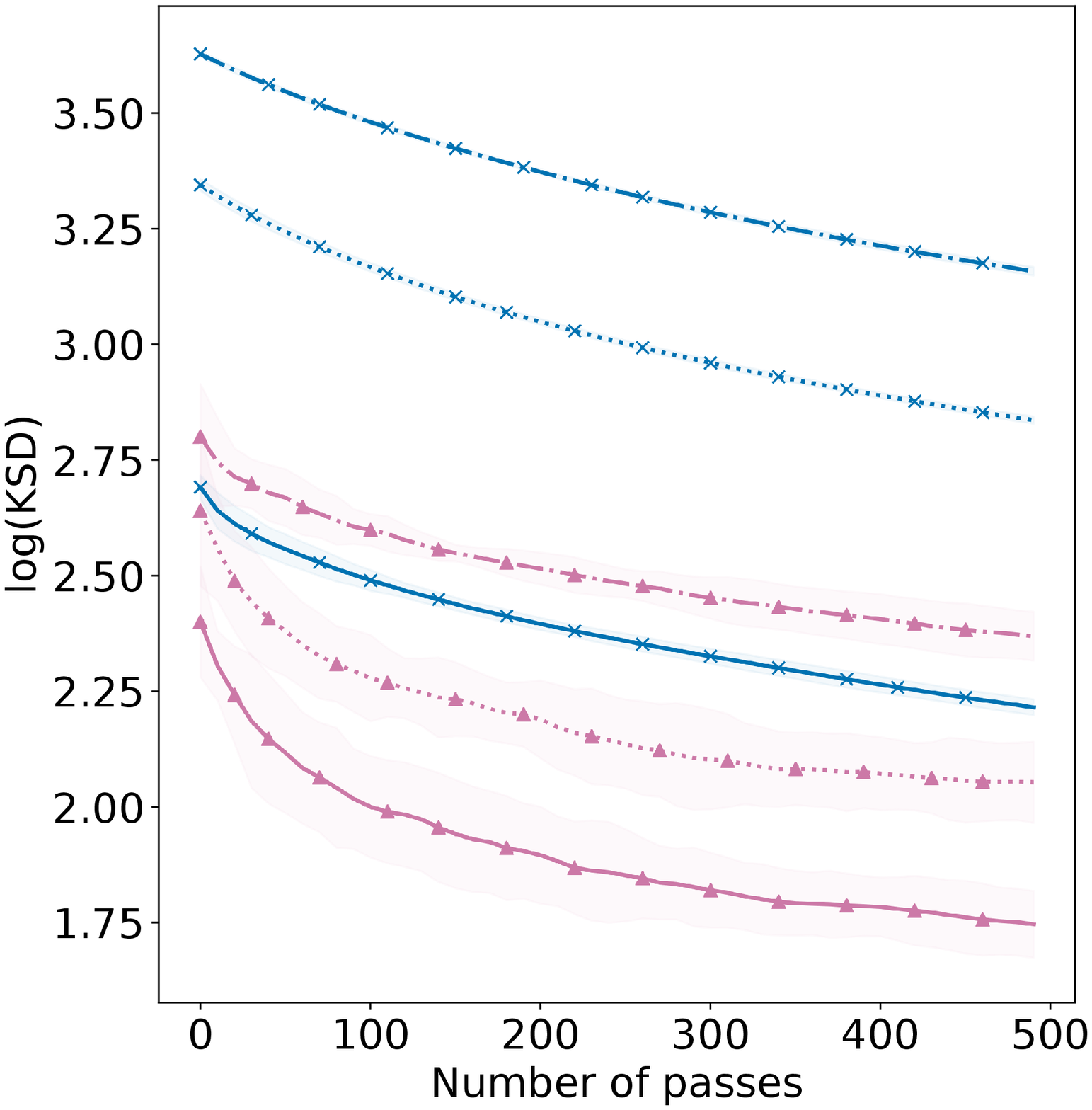} 
\vspace{-10pt}
\end{minipage} 
(b)
\end{minipage} \hspace{15pt}
\caption{Sampler performance of SGLD and SGLD-PS for 1\%, 5\% and 10\% subsample sizes over 500 passes through the data. (a) bivariate Gaussian model on synthetic data of size $N = 10^4$ (y-axis: KL divergence); (b) logistic regression on the covertype data (y-axis: KSD).}
\label{fig:sgld_comp}
\end{center} \vspace{-6pt}
\end{figure*}

\begin{figure}[h]
\begin{center}
\begin{minipage}[c]{.33\textwidth}
\centering 
\begin{minipage}[c]{\textwidth}
  \centering
\includegraphics[width=\textwidth]{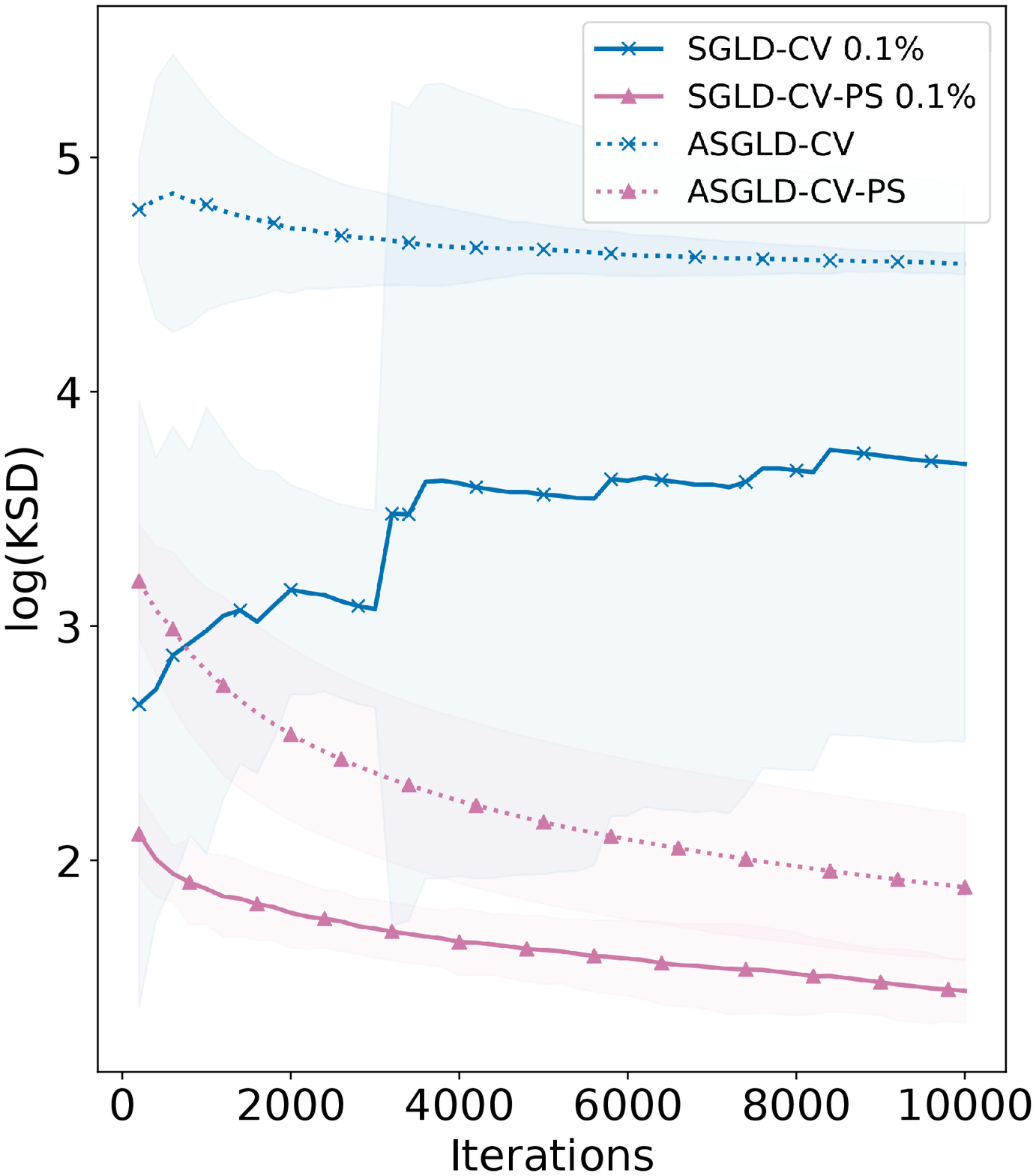} 
\vspace{-10pt}
\end{minipage}
(a) 
\end{minipage} 
\begin{minipage}[c]{.33\textwidth}
\centering 
\begin{minipage}[c]{\textwidth}
  \centering
\includegraphics[width=\textwidth]{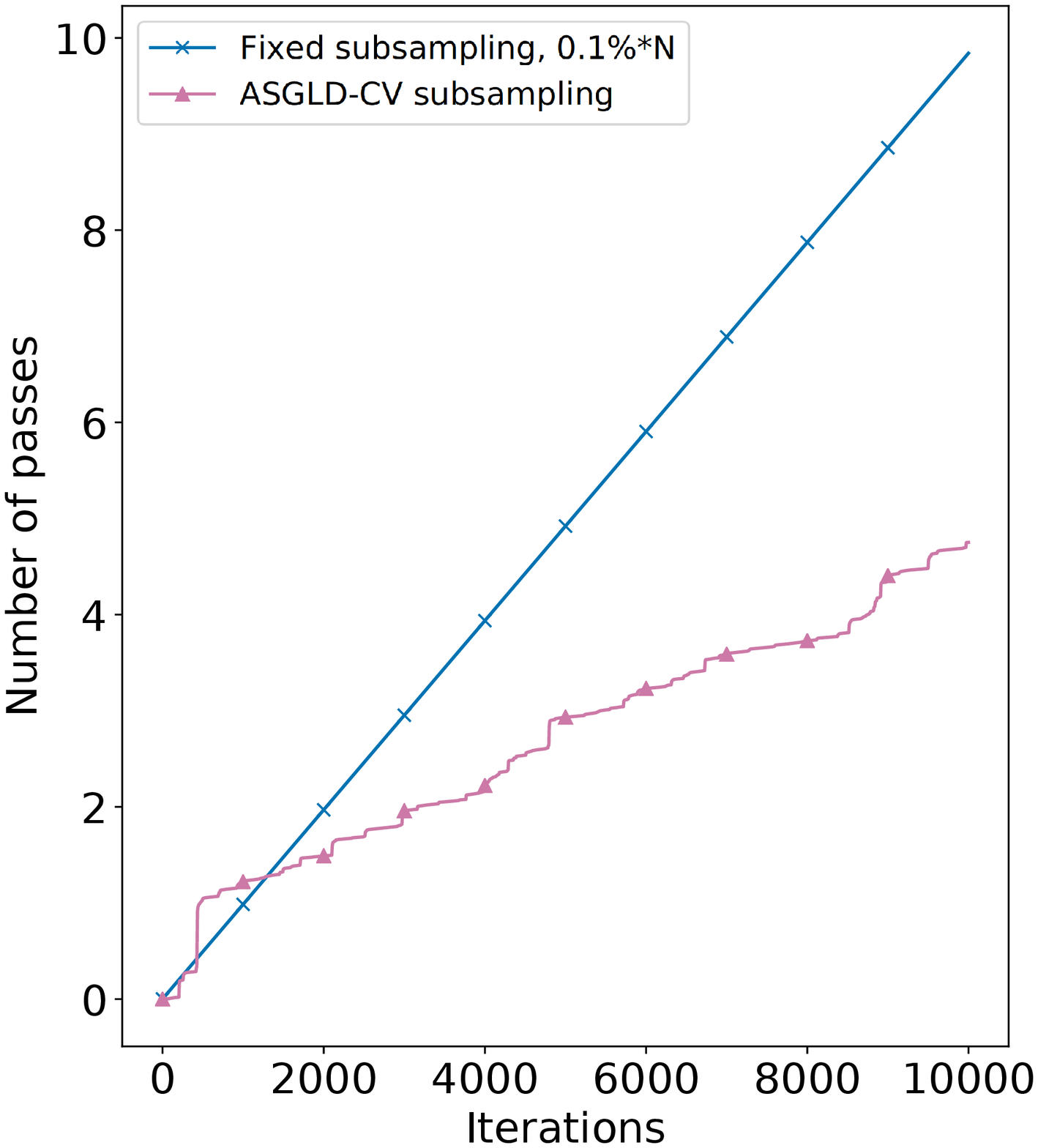}
\vspace{-10pt}
\end{minipage}
(b)
\end{minipage} \hspace{-10pt}
\begin{minipage}[c]{.33\textwidth}
\centering 
\begin{minipage}[c]{\textwidth}
  \centering
\includegraphics[width=\textwidth]{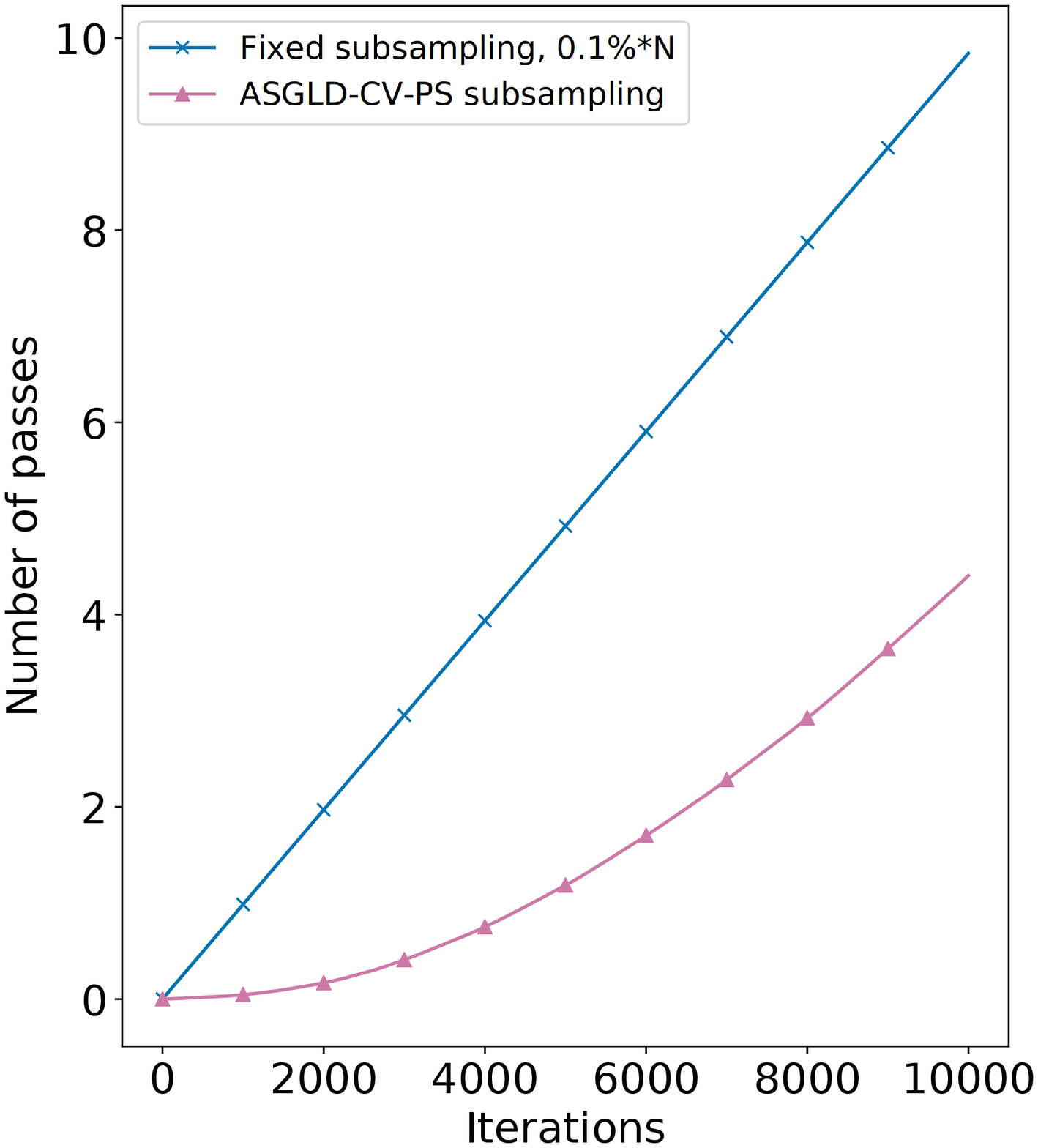} 
\vspace{-10pt}
\end{minipage}
(c)
\end{minipage} \hspace{-10pt}
\caption{The linear regression model fitted on the CASP data. (a) KSD comparison of SGLD-CV, SGLD-CV-PS, ASGLD-CV and ASGLD-CV-PS over $10^4$ iterations; (b) the number of passes through the data achieved by fixed subsampling versus ASGLD-CV; (c) the number of passes through the data achieved by fixed subsampling versus ASGLD-CV-PS.}
\label{fig:asgldcv_casp}
\end{center}
\end{figure}

SGLD-type algorithms do not mix well when the step-size is decreased to zero. It is therefore common (and in practice easier) to implement SGLD with a fixed step-size, as suggested by \cite{vollmer2016non}. For Figures \ref{fig:sgldcv_comp} - \ref{fig:asgldcv_casp}, all samplers were run with the same step-size (with $\epsilon \approx \frac{1}{N}$). This allowed us to control for discretisation error and to independently assess
the performance benefits offered by preferential subsampling. We list the step-sizes used for each experiment in  Table~\ref{tab:step-sizes} below.
\begin{table}[H]
\caption{Step-size selection} \label{tab:step-sizes} \vspace{5pt}
\begin{center}
\begin{tabular}{llll}
\textbf{FIGURE} &\textbf{DATA} &\textbf{SIZE OF DATA} &\textbf{STEP-SIZE} \\
\hline \\ 
\ref{fig:sgld_comp}(a) &synthetic bivariate Gaussian &$10,000$ &$1 \times 10^{-4}$ \\
\ref{fig:sgld_comp}(b) &synthetic balanced logistic regression &$10,000$ &$1 \times 10^{-4}$ \\
\ref{fig:sgldcv_comp}(b)-(c), \ref{fig:asgldcv_synth}  &covertype &$581,012$ &$1 \times 10^{-6}$ \\
\ref{fig:sgldcv_comp}(a), \ref{fig:asgldcv_casp} &CASP &$45,730$ &$1 \times 10^{-5}$ \\
\end{tabular}
\end{center}
\end{table}

\subsection{Initialisation}
Throughout our experiments, we were consistent in how we picked our initial start values, $\theta^{(0)}$, for SGLD and the ADAM optimiser. We sampled $\theta^{(0)}$ from the prior for the bivariate Gaussian model. Whereas, we set $\theta^{(0)} = \mathbf{0}$ for the linear and logistic regression models.

%%%%%%%%%%%%%%%%%%%%%%%%%%%%%%%%%%%%%%%%%%%%%%%%%
\section{Additional experiments}

\label{sec:app_add_exp}

\subsection{Performance comparison of SGLD and SGLD-PS}
%We consider the sampler performance of SGLD and SGLD-PS for subsample sizes of $1\%$, $5\%$ and $10\%$ of the dataset size, over 500 passes of the data. For each subsample size, we have run ten MCMC chains allowing for an equal number of iterations for both burn-in and sampling. 

%First, we consider the bivariate Gaussian model fitted on synthetic data of size $N=10^4$. Figure~\ref{fig:sgld_comp} plots KL divergence against number of passes through the data. In using the KL divergence, we are explicitly comparing the samples against the true posterior. For both methods, it is clear that larger subsample sizes allow us to converge faster to a stationary distribution that is very close to the target posterior. Overall, the KL divergence for SGLD-PS is consistently lower than for SGLD.

%Next, we consider the logistic regression model fitted on the covertype dataset \citep{blackard1998}. Figure~\ref{fig:sgld_comp}(b) plots the KSD against the number of passes through the data. In using the KSD, we are evaluating sample quality whilst taking into account the bias of each method. Overall, we can see that the KSD improves when MCMC chains are longer, irrespective of the subsample size used. We can see that the SGLD-PS samples are of better quality than those of SGLD.

\clearpage
\subsection{Performance of adaptive subsampling}
\vfill

\end{document}